\documentclass[lettersize,journal]{IEEEtran}

\usepackage{amsmath,amsfonts}
\usepackage{algorithmic}
\usepackage{algorithm}
\usepackage{array}
\usepackage[caption=false,font=normalsize,labelfont=sf,textfont=sf]{subfig}
\usepackage{textcomp}
\usepackage{stfloats}
\usepackage{url}
\usepackage{verbatim}
\usepackage{graphicx}
\usepackage{amsmath,amssymb,amsthm}
\usepackage[colorlinks,citecolor=blue]{hyperref}

\usepackage{extarrows}
\usepackage{bm}
\usepackage{xcolor}
\usepackage{tabularx}
\usepackage{multirow}
\usepackage{booktabs}
\usepackage{makecell,rotating}
 \usepackage{framed}

\begin{document}

\title{
Attribution Explanations for Deep Neural Networks: A Theoretical Perspective}

\author{Huiqi Deng, Hongbin Pei$^{\ast}$\thanks{*Corresponding author.}, Quanshi Zhang, and Mengnan Du
\IEEEcompsocitemizethanks{\IEEEcompsocthanksitem Huiqi Deng is with the School of Computer Science and Technology, Xi'an Jiaotong University, Xi'an 710049, China (email: denghq7@xjtu.edu.cn).
\IEEEcompsocthanksitem Hongbin Pei is with the School of Cyber Science and Engineering, Xi'an Jiaotong University, Xi'an 710049, China (email: peihongbin@xjtu.edu.cn).
\IEEEcompsocthanksitem Quanshi Zhang is with the Department of Computer
Science and Engineering, the John Hopcroft Center, Shanghai
Jiao Tong University, China (email: zqs1022@sjtu.edu.cn).
\IEEEcompsocthanksitem Mengnan Du is with the Department of Data Science, New Jersey Institute of Technology, NJ 07102, USA (email: mengnan.du@njit.edu).
\IEEEcompsocthanksitem This work has been submitted to the lEEE for possible publication. Copyright may betransferred without notice, after which this version may no longer be accessible.
}
}


\markboth{IEEE TRANSACTIONS ON PATTERN ANALYSIS AND MACHINE INTELLIGENCE,~Vol.~ , No.~ , 2025}%
{Shell \MakeLowercase{\textit{et al.}}: A Sample Article Using IEEEtran.cls for IEEE Journals}


\maketitle

\begin{abstract}
Attribution explanation is a typical approach for explaining deep neural networks (DNNs), which infers an attribution, importance, or contribution score of each input variable to the final network output. 
In recent years, numerous attribution methods have been developed to explain various DNNs. 
However, a persistent and fundamental concern in attribution research remains unresolved—namely, \textit{whether and which attribution methods faithfully reflect the actual contribution of input variables to the model's decision-making process}. 
This faithfulness issue significantly undermines the reliability and practical utility of attribution explanations.
We argue that these concerns primarily stem from three core challenges.
First, difficulties arise in uniformly comparing attribution methods due to their unstructured heterogeneity—significant differences in heuristics, formulations, and implementations that lack a unified organization.
Second, most attribution methods lack solid theoretical  underpinnings, with their rationales remaining largely absent, ambiguous, or unverified; 
Third, empirically evaluating faithfulness is notoriously challenging in the absence of ground truth.

Recent theoretical advances in attribution explanations provide a promising way to tackle the above challenges, 
and have attracted increasing attentions.
In this survey, we provide a comprehensive summary of these developments, with a particular emphasis on three key directions:
(i) \textit{Theoretical unification}, which uncovers key commonalities and differences among attribution methods, thereby enabling systematic comparisons;
(ii) \textit{Theoretical rationale}, which clarifies the theoretical foundations underlying existing attribution methods;
(iii) \textit{Theoretical evaluation}, which rigorously proves whether attribution methods satisfy established faithfulness principles.
Beyond a comprehensive review, we provide insights into how these studies help to deepen theoretical understanding, inform method selection, and inspire new attribution methods.
We conclude with a discussion of promising open problems for further work.
\end{abstract}


\section{Introduction}

\IEEEPARstart{O}{ver} the past decade, deep neural networks (DNNs) have shown remarkable success in various applications, particularly in computer vision, natural language processing, and intelligent decision systems. 
However, DNNs are often viewed as “black boxes,” limiting their trustworthiness—especially in high-stakes or ethically sensitive domains such as autonomous driving \cite{dikmen2016autonomous}, healthcare \cite{panesar2019machine}, and legal assistance \cite{lai2024large}.
To address this challenge, DNN explanation has received growing attention in recent years \cite{du2019techniques,murdoch2019definitions,wang2023generalized,zhao2024explainability}. 
The goal of DNN explanation is to extract and translate information about DNNs, such as structure, behavior, and mechanism, into understandable statements to humans.

\begin{figure}[t]\centering
\vspace{1mm}
\includegraphics[width = 0.49 \textwidth]{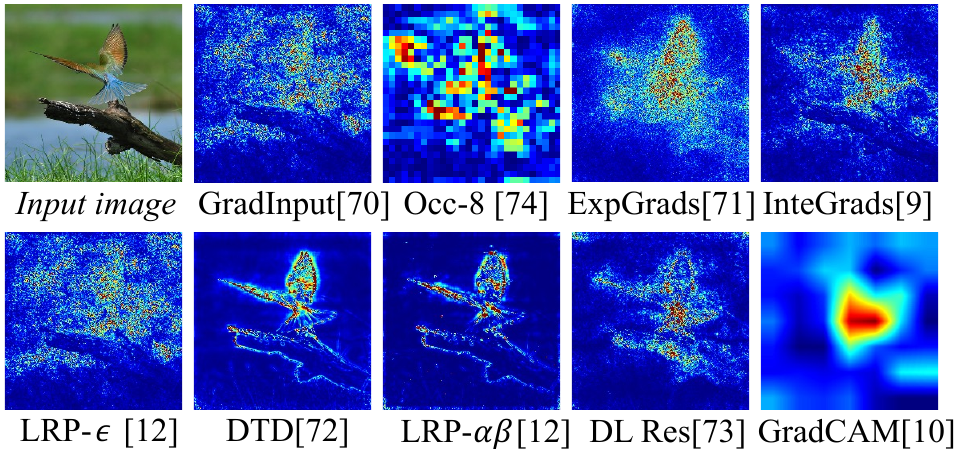}
\vspace{-4.5mm}
\caption{
An input image and the corresponding saliency maps produced by nine popular attribution methods. Each map highlights pixels deemed important for classifying the image as a ``bird''.
Notably, the attribution results vary significantly across methods.
}
\vspace{-5pt}
\label{attribution maps}
\end{figure}

Attribution explanation has emerged as a mainstream approach for interpreting DNNs \cite{baehrens2010explain, sundararajan2017axiomatic, selvaraju2017grad, springenberg2014striving, bach2015pixel, zeiler2014visualizing, lundberg2017unified, ribeiro2016should, fong2017interpretable, kindermans2017learning, nam2024illuminating}, which estimate the contribution of each input variable (e.g., a pixel in an image or a word in a sentence) to the final output of a DNN.
Mathematically, given a pretrained DNN $f(\cdot)$ and an $n$-dimensional input sample $\bm{x}  = [x_1,\dots x_n]^{\rm T} \in \mathbb{R}^n$, attribution explanations aim to produce an attribution vector $\bm{a} = [a_1,\dots, a_n ]^{\rm T} \in \mathbb{R}^n$, where $a_i$ reflects $x_i$'s  influence on the model output $f(\bm{x}) \in \mathbb{R}$.
In image tasks, these scores can be visualized as saliency maps, offering intuitive visual explanations.
As shown in Fig.~\ref{attribution maps}, nine representative attribution methods produce saliency maps that highlight pixel-wise contributions to a bird classification task.

Numerous attribution methods have been proposed recently and widely applied to understanding various DNNs, including cutting-edge models such as vision transformers (ViTs) for image classification \cite{selvaraju2017grad,covert2023learning}, 
diffusion models for image generation \cite{tang2023daam}, and
large language models (LLMs) \cite{kokalj2021bert, li2023survey}. 
Moreover, attribution methods  are also leveraged in various other applications, such as automated model debugging \cite{lertvittayakumjorn2021explanation, anders2022finding},  guiding model design
\cite{erion2021improving, rieger2020interpretations}, 
continuously refreshing the information base of LLMs \cite{li2023survey},  
inspiring mathematical conjectures \cite{davies2021advancing}, 
exploring molecular structure-activity relationships in chemistry  \cite{wu2023chemistry}, and mitigating shortcut bias in AI-based COVID-19 detection \cite{degrave2021ai}.

Despite these widespread applications, attribution methods face a core faithfulness concern \cite{doshi2017towards, lipton2018mythos, kindermans2019reliability, leavitt2020towards, srinivas2020rethinking}:
\textit{do their explanations faithfully reflect the actual contribution of input variables to model decisions?}
This concern has sparked growing debate over the reliability of attribution explanations.
Faithfulness is not just a desirable property; it is a foundational requirement for a trustworthy explanation. 
This becomes especially critical in high-stakes applications, where misleading attributions may distort scientific conclusions, obscure medical diagnoses, or reinforce social bias \cite{rudin2019stop, ghassemi2021false}.
Essentially, these concerns arise from three fundamental issues. 
\begin{itemize}
\item \textbf{Lack of unified foundations hinders systematic comparison}.
Existing attribution methods are rooted in diverse heuristic motivations, mathematical formulations, and implementation strategies \cite{ancona2017towards, 10414149}.
Consequently, their generated attribution results often differ significantly, as illustrated in Fig.~\ref{attribution maps}.
These discrepancies, in the absence of a unified theoretical framework,  make it difficult to uniformly understand or compare attribution methods.
Moreover, such inconsistencies  suggest that some methods inevitably lack faithfulness, assuming the uniqueness of a truly faithful attribution.
\vspace{2pt}

\item \textbf{Lack of theoretical rationales undermines guarantees of faithfulness}. 
Most attribution methods are developed heuristically, with limited or no formal theoretical justification \cite{samek2021explaining}.
Their underlying rationales often remain unclear, unverified, or absent.
Consequently, the faithfulness of these methods cannot be rigorously 
guaranteed from a theoretical perspective.
\vspace{2pt}

\item \textbf{Challenges on empirical evaluation hinder assessing faithfulness}. The absence of ground truth attribution presents a significant challenge in empirically evaluating attribution faithfulness.
To address the challenge, many alternative evaluation strategies are proposed \cite{yang2019evaluating,yang2019bim,schulz2020restricting,rao2022towards,rong2022consistent,zhou2022feature,ju2022logic,arias2023confusion, gevaert2024evaluating},
such as synthesizing a dataset with ground truth attributions \cite{yang2019bim} and assessing alignments between attribution results and human cognition \cite{schulz2020restricting}.
However, none of them is widely accepted as objective \cite{zhou2022feature, sithakoul2024beexai}. 
Moreover, different strategies often present very different evaluation conclusions \cite{tomsett2020sanity, duan2025evaluation}.
In summary, empirical evaluations alone cannot resolve faithfulness disputes.
\end{itemize}

\noindent\textbf{Theoretical advances as promising solution.}
Recent advancements in theoretical studies of attribution explanations provide a promising way to tackle these issues discussed above. 
A notable example is the \textit{Shapley value}, which has been shown to be the unique solution satisfying a set of axiomatic faithfulness principles \cite{lundberg2017unified},  
thereby providing strong theoretical justification for its attribution faithfulness and widespread adoption \cite{sundararajan2020many, merrick2020explanation, chen2022explaining, kwon2022weightedshap}.   
Another influential example  is provided by Nie et al. \cite{nie2018theoretical}, who theoretically prove that \textit{Deconv} and \textit{GBP} methods do not reflect the DNN's decision-making process; instead, they only recover input images.
Consequently, this prompts the community to  critically reconsider the faithfulness of these methods and to use them with greater caution \cite{murdoch2019definitions, adebayo2018sanity,  lyu2024towards}.
In summary, these theoretical  approaches aim to provide principled and generalizable criteria for understanding and assessing attribution methods.

\begin{figure*}[t]\centering
\includegraphics[width = 0.95 \textwidth]{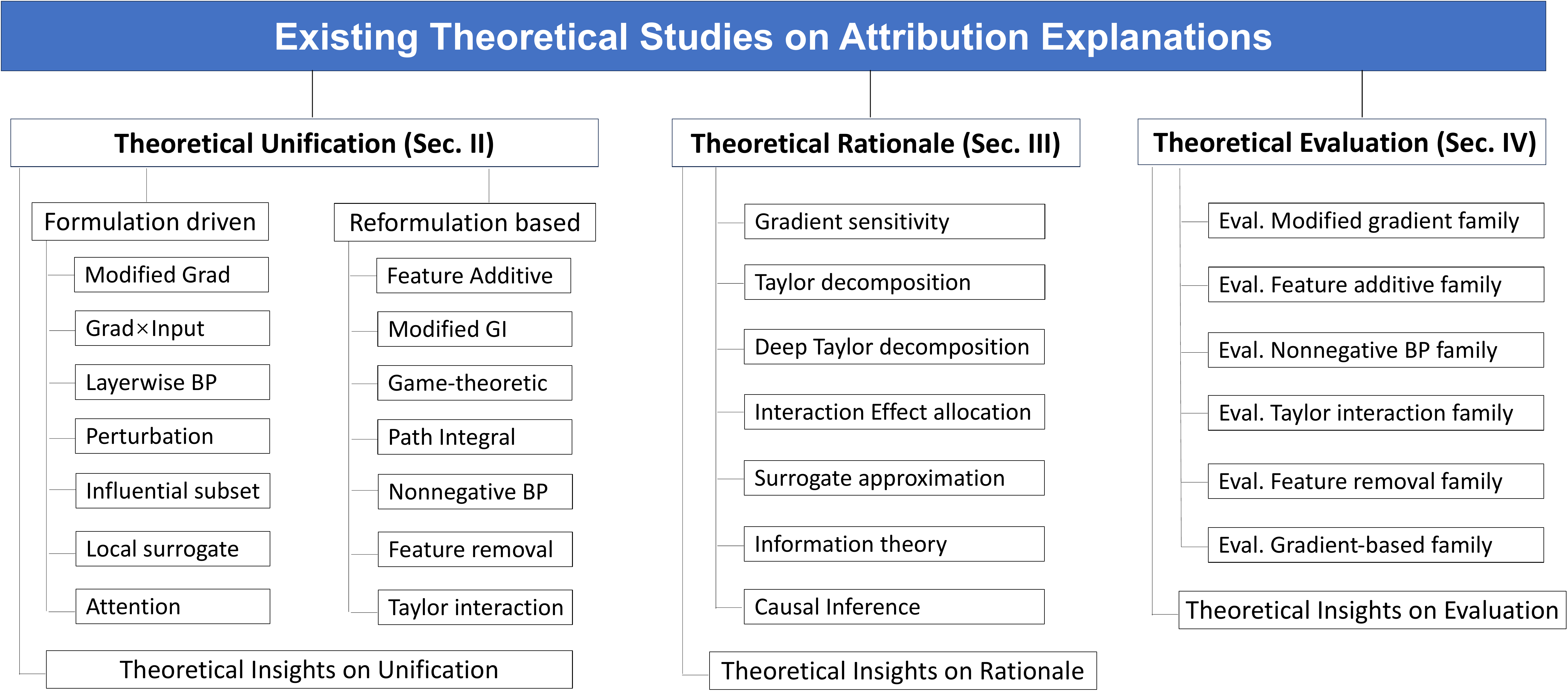}
\vspace{-2pt}
\caption{Overview of existing theoretical studies on attribution explanations. 
The diagram summarizes three core dimensions: theoretical unification (Sec.~\ref{Sec:theoretical unification}), theoretical rationale (Sec.~\ref{Sec:theoretical rationale}), and theoretical evaluation (Sec.~\ref{Sec:theoretical evaluation and comparison}), along with our corresponding insights at the end of each section.}
\label{Overview}
\end{figure*}

To synthesize these growing theoretical insights, this survey presents a comprehensive and structured review of recent theoretical  developments in attribution research.
To better organize these diverse efforts, we introduce a taxonomy that aligns theoretical studies with the three major challenges discussed earlier.
As illustrated in Fig.~\ref{Overview}, these studies fall into the following three core dimensions:
\begin{itemize}
\item \textbf{Theoretical unification}, which aims to unify diverse attribution methods under a common framework, where each method corresponds to a specific form of a shared formulation.
It clarifies key commonalities across seemingly distinct methods.
For example, methods like \textit{LRP-$\epsilon$}, \textit{Grad$\times$Input}, \textit{DeepLIFT}, and \textit{Integrated Gradients} can be unified under a modified gradient-based formulation, highlighting their core similarities \cite{ancona2017towards}.
\vspace{1pt}

\item \textbf{Theoretical rationale}, which aims to clarify the underlying mechanisms that justify the use of each attribution method. 
This helps clarify the extent to which a method is supported by sound theoretical foundations.
For example, from a causal inference perspective,  the seemingly heuristic \textit{Occ-1} method can be interpreted as a special case of the individual causal effect (ICE) \cite{chattopadhyay2019neural}.
\vspace{1pt}

\item \textbf{Theoretical evaluation}, which seeks to rigorously assess whether attribution methods satisfy established faithfulness principles or robustness properties through formal analysis and theoretical proof.
For example, it has been formally proven that many attribution methods violate fundamental  faithfulness principles such as output sensitivity and parameter sensitivity \cite{sixt2020explanations}.
\end{itemize}
Together, three dimensions provide a cohesive and mutually reinforcing framework for clarifying, justifying, and evaluating attribution methods from a theoretical standpoint.
First, unification offers a structured foundation for analyzing rationales and assessing faithfulness, by revealing shared formulations across methods.
Rather than focusing on isolated methods, typical theoretical studies are conducted at the level of  attribution families, where a unified formulation enables generalized proofs, broader insights, and more robust comparisons.
Moreover, investigating  rationales offers principled justifications for faithfulness and can itself be regarded as a key component of theoretical evaluation. 
In this sense, rationales complement and enrich the broader scope of theoretical evaluation.

\begin{figure}[t]\centering
\includegraphics[width = 0.475 \textwidth]{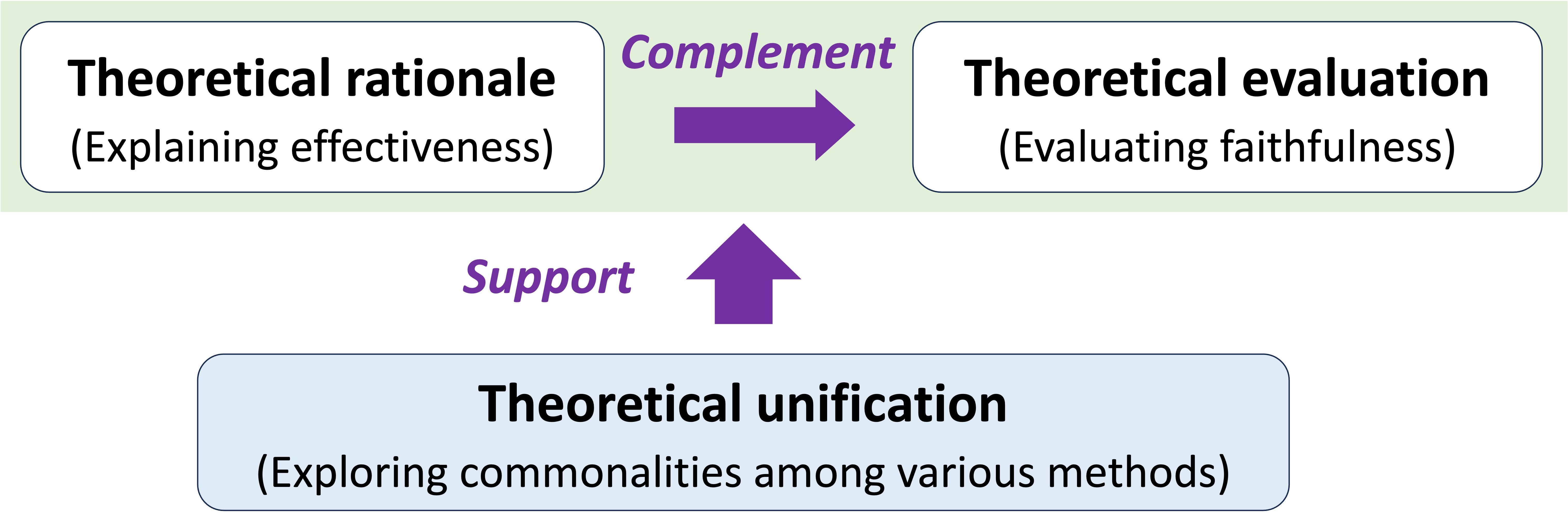}
\vspace{-3pt}
\caption{Connections among three theoretical dimensions: \textit{unification} reveals shared structures, serving as the foundation; \textit{rationales} explain why methods are effective; and \textit{evaluation} assesses whether methods are faithful. 
Rationale analysis complements evaluation by providing principled justification.}
\label{three_dimension}
\vspace{-5pt}
\end{figure}

Beyond a comprehensive review, we further provide insight into how recent theoretical studies contribute to a deeper understanding of attribution methods, provide guidance for method selection, and inspire new attribution techniques and evaluation frameworks.
\vspace{2pt}

\noindent \textbf{Novelty and contributions}.
While existing surveys on attribution explanations primarily focus on methodological taxonomies \cite{zhang2018visual,linardatos2020explainable,zhang2021survey,li2022interpretable,madsen2022post,yuan2022explainability} or empirical evaluation protocols \cite{lyu2024towards,li2022interpretable,carvalho2019machine,zhou2021evaluating,chan2022comparative,nauta2023anecdotal}, theoretical advancements remain fragmented and lack systematic organization.
In contrast, \textit{this survey bridges this gap by providing the first structured and comprehensive investigation of the theoretical progress in attribution explanations}. 
This theoretical perspective is particularly timely and important for attribution research, as the persistent challenge of empirically unverifiable faithfulness underscores the urgent need for rigorous theoretical foundations.

Specifically, our key contributions are as follows:
(1) We offer a comprehensive, well-structured, and systematic review of fragmented theoretical attribution research, organizing them into three interrelated dimensions. 
(2) We present an integrative perspective on how theoretical studies deepen the understanding of attribution methods—particularly their faithfulness, provide principled guidance for method design and usage, and inspire the development of new techniques.
\vspace{2pt}

\noindent\textbf{Organization}.
This paper is organized as follows. As shown in Figure~\ref{Overview}, 
Section~\ref{Sec:theoretical unification}-~\ref{Sec:theoretical evaluation and comparison}
comprehensively review existing work on theoretical unification, theoretical rationale, and theoretical evaluation, respectively. 
Section ~\ref{subsec:our perspective1},~\ref{subsec:our perspective2}, and~\ref{subsec:our perspective3}, present our insights into how these three dimensions contribute to a deeper theoretical understanding.
Section~\ref{sec:Expected benefits}  illustrates how these theoretical advances can inform the practical use and development of attribution methods.
Finally, section~\ref{sec:Future work} outlines future  directions for theoretical developments.

\section{Theoretical unification}\label{Sec:theoretical unification}
Existing attribution methods are typically grounded in various heuristic, mathematical formulations, and implementation details, often resulting in significantly different attribution results, as shown in Fig. \ref{attribution maps}. 
However, the theoretical connections between these methods—particularly their key commonalities and differences—remain unclear. This lack of clarity makes it challenging to systematically understand and evaluate these methods from a theoretical perspective.

To address this issue, several studies have attempted to theoretically unify attribution methods by uncovering the key commonalities among them. 
To synthesize these efforts, we adopt a dual-perspective organizational scheme that reflects both the historical design philosophy and deeper theoretical alignments across methods.  
Specifically, we employ two complementary paradigms:
\begin{itemize}
\item \textbf{Formulation-Driven Unification} 
(Sec \ref{subsec: Formulation-based unification}).
This perspective categorizes attribution methods into standard families, each defined by a representative mathematical formulation that embodies the method's original design philosophy.
It offers a widely recognized and intuitive taxonomy, clarifying how various methods were independently developed.
\vspace{2pt}

\item \textbf{Reformulation-Based Unification} (Sec \ref{subsec: Other unification framework}).
Rather than following historical design logic, this perspective identifies theoretical unifications by re-deriving and aligning mathematical expressions across different methods.
Such reformulations uncover deeper intrinsic connections among methods from different standard families, offering a more enriched understanding of attribution methods.
\end{itemize}

Beyond reviewing existing research, Section~\ref{subsec:our perspective1} presents our insights into how theoretical unification contributes to a deeper understanding of attribution methods.

\begin{table}[t]
\caption{Main symbols and terminologies in this paper.}
\label{Tab:Basic symbols}
\renewcommand\arraystretch{1.35}
\centering
\begin{tabular}{>{\centering}p{17.5mm} | p{52.5mm} }
\toprule
\textbf{{\makebox[0.06\textwidth]{Symbol} }} &  \textbf{Description}  
\\  \hline
$f$ &  a pre-trained DNN to be explained \\ \hline
$\bm{x}$ &  $\bm{x} = [x_1,\dots, x_n]^{\rm T}$, input sample  \\ \hline
$x_i$ & the $i$-th input variable \\ \hline
$f(\bm{x})$ &  network output on the input sample $\bm{x}$  \\ \hline
$\bm{a}$ &  $\bm{a} =  [a_1,\dots, a_n ]^{\rm T}$, attribution vector  \\ \hline
$b_i$ &  baseline value to mask variable $x_i$  \\ \hline
$\bm{b}$ &  $\bm{b} = [b_1,\dots, b_n]^{\rm T}$, baseline sample \\ \hline
$N$ &  $[1, \dots, n]^T$, index set of input variables  \\ \hline
$S$ &  $S \subseteq N$, subset of $N$   \\ \hline
$\bar S$ &  complementary set  of $S$   \\ \hline
$f(\bm{x}_S)$ &  output when variables in $\bar S$ are masked  \\ \hline
$\bm{y}$ & $\bm{y} =  [f^1(\bm{x}),\dots, f^C(\bm{x})]^{\rm T}$,  output vector \\ \hline
$\bm{x}^{(l)}$ &  features in the $l$-th layer of DNN \\ \hline
$\bm{a}^{(l)}$ &  attribution of features $\bm{x}^{(l)}$ \\ \hline
$M^{(l)}$&  BP matrix for attributions $\bm{a}^{(l)}$ \\ \hline
$M^{(l), +}$ & nonnegative BP matrix for attributions \\ \hline
$g$ &  surrogate model of the DNN $f$\\ \hline
$\partial^{\mathbb{M}} f /\partial x_i$ &  modified gradient with designed BP rules \\ \hline
$\phi(i)$ &  independent effect of individual variable $i$ \\ \hline
$I(S)$ &   interaction effect between variables in $S$ \\ 
\bottomrule
\end{tabular}
\vspace{- 3 pt}
\end{table}

\subsection{Formulation-Driven Unification}
\label{subsec: Formulation-based unification}
Formulation-Driven unification categorizes  existing mainstream attribution methods into seven distinct and orthogonal families, based on their core mathematical formulations that reflect the original design philosophies of these methods.
Each attribution family is characterized by a representative equation,  which reveals the shared commonalities among methods, as summarized in Table \ref{tab:Formulation-based unifications}.
\vspace{2pt}

\textbf{Contribution Highlight.} Although the formulation-driven categorization has been widely used in practice, formal mathematical representations for these categories are often absent or underdeveloped in the literature.
We address this gap by systematically deriving \textit{explicit unified formulations} for each attribution family, thereby enhancing the theoretical rigor of attribution unification.
Moreover, these formulations allow us to pinpoint the \textit{key differences} among methods within the same family, offering a more precise understanding on their intra-family variations. \\

\begin{table*}[t]
\caption{Unification of Formulation-Driven Attribution Families  \\
(Unified formulations, representative methods, and key difference)}
\label{tab:Formulation-based unifications}
\renewcommand\arraystretch{2.5}
\centering
\begin{tabular}
{>{\centering} p{23mm} |   >{\raggedright} p{65mm} | >{\centering\arraybackslash} p{55mm} | >{\centering\arraybackslash} p{23mm}  }
\toprule
\textbf{Attribution  Family}
& \makecell[c]{\textbf{Unification  Formulation}}
& \makecell[c]{\textbf{Representative Attribution Methods}}
& \makecell[c]{\textbf{Key Difference}}
  \\ \hline 
  
Modified gradient
&  $ \displaystyle a_i = \partial^{\mathbb{M}} f(\bm{x}) / \partial x_i 
\ (\textcolor{gray}{Eq. ~\ref{eqn:Modified gradient based attribution family})}$
&  \textit{Grad} \cite{baehrens2010explain}, 
 \textit{SG} \cite{smilkov2017smoothgrad}, \textit{Deconv} \cite{zeiler2014visualizing},    \textit{GBP} \cite{springenberg2014striving}
&  Def. of  $\frac{\partial^{\mathbb{M}} f(\bm{x})}{\partial x_i}$
 \\ \hline
 
Gradient $\times$ Input
& $\displaystyle a_i = \mathbb{E} [ \frac{\partial f(\bm{x})}{\partial x_i} ] \cdot (x_i - b_i) \ (\textcolor{gray}{Eq. ~\ref{eqn:GradientInput based attribution family})}$
& \textit{Grad$\times$Input} \cite{simonyan2013deep},
\textit{IG} \cite{sundararajan2017axiomatic}, 
\textit{EG} \cite{erion2019learning}, 
\newline 
\textit{GradCAM} \cite{selvaraju2017grad} 
& baseline $\bm{b}$, \quad \quad
avg. scope 
\\  \hline 

Layerwise  BP
&  $\displaystyle \bm{a}  = \prod_{l = 1}^L M^{(l)} \bm{y} \
(\textcolor{gray}{Eq. ~\ref{eqn:layer-wise backpropagation based attribution family})}$
&   \textit{LRP-$0$/$\epsilon$/$\alpha\beta$} \cite{bach2015pixel},  \textit{DTD} \cite{montavon2017explaining},
\textit{DL-Res/Rev} \cite{shrikumar2017learning}, 
\textit{DSHAP} \cite{lundberg2017unified}, \textit{PatternNet} \cite{kindermans2017learning}, 
$\dots$
& BP matrix \quad \quad $M^{(l)}$
\\ \hline

Perturbation
&$\displaystyle a_i = \sum_{S}  w_{S} \cdot [F(\bm{x}_{S \cup \{i\}}) - F(\bm{x}_{S})] 
 \ (\textcolor{gray}{Eq. ~\ref{eqn:Perturbation-based attribution family})}$
& \textit{Occ-1}  \cite{zeiler2014visualizing}, \textit{Occ-p} \cite{zintgraf2017visualizing}, 
\textit{PDiff} \cite{zintgraf2017visualizing}, \textit{Shapley}  \cite{lundberg2017unified}, 
\textit{Banzhaf} \cite{dubey1979mathematical},  \textit{RISE} \cite{Petsiuk2018rise}, 
\textit{ACE} \cite{chattopadhyay2019neural}, ...
& Def. of $F(\bm{x}_{S})$, 
 weight $w_S$
 \\ \hline

Influential subset
& $\displaystyle S^* = \arg\min_S f(\bm{x}_{\bar S}) + \lambda_1 |S| + \lambda_2 R(S) (\textcolor{gray}{ Eq. ~\ref{eqn: most-influential subset})}$
& \textit{MP} \cite{fong2017interpretable},  \textit{EP} \cite{fong2019understanding}, 
\textit{RTIS}  \cite{dabkowski2017real}, \textit{IB} \cite{schulz2020restricting}
& regularizer $R(S)$ 
\\ \hline

Local surrogate
& $\displaystyle \bm{a} = extract\left(\arg\min_{g \in \mathcal{G} } \mathcal{L}(f, g, \mathcal{N}_{\bm{x}})+ \mathcal{C}(g)\right) (\textcolor{gray}{Eq. ~\ref{eqn:Local surrogate based attribution family})}$
& \textit{LIME}  \cite{ribeiro2016should},  \textit{OptiLIME} \cite{visani2020optilime}, 
 \textit{S-LIME}  \cite{zhou2021s}, 
 \newline
 \textit{BayLIME} \cite{zhao2021baylime} 
& sampling stability strategy for $\mathcal{N}_{\bm{x}}$
 \\ \hline

Attention
& $\displaystyle \bm{a} = Aggregate (W_A) \ (\textcolor{gray}{Eq. ~\ref{eqn:Attention based attribution family})}$
&\textit{Self-attention} \cite{hao2021self}, \textit{DAAM} \cite{tang2023daam}
& aggregator $Aggregate(\cdot)$
\\ 
\bottomrule
\end{tabular}
\end{table*}

\noindent \textbf{(1) Modified Gradient Attribution Family}.
It is widely acknowledged that the gradient $\partial f(\bm{x})/\partial x_i$ of a model's output with respect to an input variable can serve as an indicator of the variable’s relative importance.
Methods in the modified gradient attribution family adopt a gradient-based formulation and compute attribution as follows:
\begin{equation}\label{eqn:Modified gradient based attribution family}
a_i = \partial^{\mathbb{M}} f(\bm{x}) / \partial x_i 
\end{equation}
Here, $\partial^{\mathbb{M}} f(\bm{x}) / \partial x_i$ denotes a modified gradient computed using specific backpropagation rules.

Representative methods in this family include \textit{Gradient (Grad)} \cite{baehrens2010explain}, \textit{Smooth gradients (SG)} \cite{smilkov2017smoothgrad},  \textit{Deconv} \cite{zeiler2014visualizing}, and \textit{Guided Back-Propagation (GBP)} \cite{springenberg2014striving}.
These methods \textbf{differ in} how they define and apply gradient back-propagation rules.

\vspace{6pt}
\noindent \textbf{(2) Gradient$\times$Input Attribution Family}.
Methods in this family formulate attributions as the element-wise product between input features and their corresponding gradients. The unified formulation is given by:
\begin{equation}\label{eqn:GradientInput based attribution family}
a_i = \mathbb{E} [\frac{\partial f(\bm{x})}{\partial x_i}] \cdot (x_i - b_i)
\end{equation}
where $b_i$ is a predefined baseline value to represent the masking state of $x_i$, and $\bm{b} = [b_1, \dots, b_n] \in \mathbb{R}^n$ denotes the baseline sample. 

Representative methods in this family include \textit{Gradient$\times$Input} \cite{simonyan2013deep}, \textit{Integrated gradients (IG)}  \cite{sundararajan2017axiomatic},  \textit{Expected Gradients (EG)} \cite{erion2019learning}, \textit{Grad-CAM} \cite{selvaraju2017grad, zhou2016learning}, and so on.
The \textbf{main difference} among methods in this family lies in the choice of baseline and how gradients are averaged (e.g., across input samples or along integration paths). 

\vspace{6pt}
\noindent \textbf{(3) Layer-Wise Backpropagation Attribution Family}.
This family estimates attribution for features at each intermediate layer, and  back-propagates these attributions recursively through the network. 
Formally,  attributions are propagated from the output layer to the input layer using a series of backpropagation matrices $M^{(l)}$, as follows:
\begin{equation}\label{eqn:layer-wise backpropagation based attribution family}
\begin{aligned}
\bm{a}^{(l-1)}  & = M^{(l)} \bm{a}^{(l)}, \\
\bm{a}  \overset{\rm{def}}{=} \bm{a}^{(0)} & = \prod\nolimits_{l = 1}^L M^{(l)} \bm{y}
\end{aligned}
\end{equation}
Here, $\bm{a}^{(L)} = \bm{y} \in \mathbb{R}^{n_L}$ denotes the DNN output vector. 
Each back-propagation  matrix  $M^{(l)} \in \mathbb{R}^{n_{l-1} \times n_{l}}$ governs the flow of attribution from layer $l$ to layer $l-1$, where each element $M_{i,j}^{(l)}$ represents how much the attribution of feature $j$ at layer $l$ contributes to the attribution of feature $i$ at layer $l-1$.
The final attribution at the input layer  is obtained by multiplying a series of back-propagation matrices with the output vector. 

Typical methods in this attribution family include \textit{LRP-$0$}, \textit{LRP-$\epsilon$},  \textit{LRP-$\alpha\beta$}  \cite{bach2015pixel}, \textit{DeepLIFT Res (DL-Res)},  \textit{DeepLIFT Rev (DL-Rev)} \cite{shrikumar2017learning}, \textit{Deep Taylor Decomposition (DTD)} \cite{montavon2017explaining}, \textit{PatternNet} \cite{kindermans2017learning},  \textit{Excitation BP (ExBP)} \cite{zhang2018top}, \textit{Rect Gradients (RectG)} \cite{kim2019saliency}, and \textit{Deep SHAP} \cite{lundberg2017unified}.
The \textbf{main differences} among these methods lie in the definition and construction of the backpropagation matrices $M^{(l)}$.

\vspace{6pt}
\noindent \textbf{(4) Perturbation-Based Attribution Family}.
This family infers the attribution of an input variable according to how much perturbing (or masking) the variable alters the network output. Formally, the attribution $a_i$  is formulated as the weighted average  of the output changes caused by perturbing the $i$-th variable, \textit{i.e.},
\begin{equation}\label{eqn:Perturbation-based attribution family}
\begin{aligned}
a_i = \sum\limits_{S \subseteq N \setminus \{i\}} & w_{S} \cdot [F(\bm{x}_{S \cup \{i\}}) - F(\bm{x}_{S})], \\
\text{where } \ F(\bm{x}_S) &= \mathbb{E}_{\bm{b}_{\bar S} \sim  p(\bm{b}_{\bar S})} [f(\bm{x}_S)]
\end{aligned}
\end{equation}
Here, $\bm{x}_S$ denotes a masked sample where variables in $S$ remain unchanged but variables in $\bar S$ replaced by their corresponding baseline values $\bm{b}_{\bar S}$.
Then, $F(\bm{x}_S)$ represents the expected network output for the perturbed sample $\bm{x}_S$, averaged over baseline values sampled from the distribution  $p(\bm{b}_{\bar S})$. 
In this way, the difference  $F(\bm{x}_{S \cup \{i\}}) - F(\bm{x}_{S})$ measures the marginal effect of unmasking variable $i$, with variables in $S$ serving as the context.

Typical methods in this family include  \textit{Occlusion-1 (Occ-1)} \cite{zeiler2014visualizing}, \textit{Occlusion-patch (Occ-p)} \cite{zintgraf2017visualizing}, \textit{Prediction difference (PDiff)} \cite{zintgraf2017visualizing}, \textit{Shapley value}  \cite{lundberg2017unified, strumbelj2010efficient}, \textit{SAGE} \cite{covert2020understanding}, and   \textit{Banzhaf value} \cite{dubey1979mathematical}, \textit{RISE} \cite{Petsiuk2018rise}, \textit{Average Causal Effect (ACE)} \cite{chattopadhyay2019neural}, and so on. 
The \textbf{main difference} among them lies in  the definition of model output  $F(\bm{x}_{S})$ and the weighting scheme $w_S$ over contextual subsets $S$. 

\vspace{6pt}
\noindent \textbf{(5) Influential Subset Attribution Family}.
This family aims to identify the most influential subset of input variables, which is defined as the minimal subset $S$ whose masking leads to the greatest degradation in model output.
Formally, the most influential subset $S$ is determined by solving:
\begin{equation}\label{eqn: most-influential subset}
S^* = \arg\min_S f(\bm{x}_{\bar S}) + \lambda_1 \cdot |S| + \lambda_2 \cdot R(S).
\end{equation}
where $\lambda_1, \lambda_2 > 0$ balance the sparsity term $|S|$ and theregularization term $R(S)$.

Representative methods in this family include \textit{Meaningful Perturbation (MP)}  \cite{fong2017interpretable},  \textit{Extremal Perturbation (EP)} \cite{fong2019understanding}, \textit{Real Time Image Saliency (RTIS)} \cite{dabkowski2017real}, \textit{Information Bottleneck (IB)} \cite{schulz2020restricting}, and others \cite{du2018towards, wagner2019interpretable, fu2021differentiated}.  
Due to the non-convex nature of the objective function,  these methods incorporate various regularizers to attain a more interpretable local optima.  The \textbf{main difference} among these methods lies in the choice of regularizers $R(S)$, such as total-variation \cite{fong2017interpretable} or area constraint \cite{fong2019understanding}.

\vspace{6pt}
\noindent \textbf{(6) Local Surrogate Attribution Family}.
This family approximates the local behavior of a DNN $f$ by fitting a human-understandable surrogate model $g \in \mathcal{G}$ in the neighborhood $\mathcal{N}_{\bm{x}}$ of input sample $\bm{x}$.
Attributions are extracted from the fitted surrogate: 
\begin{equation}\label{eqn:Local surrogate based attribution family}
\begin{aligned}
g & = \arg\min_{g \in \mathcal{G} }      \mathcal{L}(f, g, \mathcal{N}_{\bm{x}})+ \mathcal{C}(g),  \\
\Rightarrow \ \ \bm{a} & = extract (g)
\end{aligned}
\end{equation}
where $\mathcal{L}(f, g, \mathcal{N}_{\bm{x}})$ quantifies how well the surrogate model $g$ approximates $f$ within the local region $\mathcal{N}_{\bm{x}}$, 
and $\mathcal{C}(g)$ penalizes the model complexity of $g$.

A representative method is \textit{LIME} \cite{ribeiro2016should}, where the surrogate model is typically set as a linear model $g = \bm{w}^T \bm{x}$:
\begin{equation}
\label{eqn:LIME}
\begin{aligned}
\bm{w}^*  = \arg\min_{\bm{w}}   \sum\limits_{\bm{\tilde x} \in \mathcal{N}_{\bm{x}}} & \pi_{\bm{\tilde x}} \cdot ||f(\bm{\tilde x}) - \bm{w}^T \bm{\tilde x}||_2^2 + \lambda \cdot |\bm{w}|. \\
\end{aligned}
\end{equation}
where $\pi_{\bm{\tilde x}}$ denotes the importance of each neighbor $\bm{\tilde x}$. 
The optimized weight  $\bm{w}^*$ serves as the attribution vector.

Subsequent variants of \textit{LIME}, such as \textit{OptiLIME} \cite{visani2020optilime}, \textit{S-LIME}  \cite{zhou2021s}, and \textit{BayLIME} \cite{zhao2021baylime}, were developed to address the instability problem in \textit{LIME} explanations  caused by the random sampling from the neighbor $\mathcal{N}_{\bm{x}}$. 
The \textbf{main difference} among these methods lies in their strategies to improve the stability of \textit{LIME}.  

\vspace{6pt}
\noindent \textbf{(7) Attention Based Attribution Family}.
This family focuses on explaining DNNs that incorporate attention mechanisms, such as the widely used BERT model for NLP \cite{devlin2018bert}, vision Transformers for image classification \cite{dosovitskiy2020image} and diffusion models in vision generation \cite{ho2020denoising}. 
Attributions are typically derived by aggregating attention weight matrices $W_A$. 
\begin{equation}\label{eqn:Attention based attribution family}
\bm{a} = Aggregate (W_A)
\end{equation}

Representative  methods in this attribution family include  \textit{Self-attention} \cite{hao2021self}, \textit{DAAM} \cite{tang2023daam},
and others \cite{clark2019does, li2023does, liu2024towards}. The \textbf{main difference} among these methods lies in the strategies to interpret and aggregate the attention weights $W_A$.

\begin{table*}[t]
\caption{Unification of Reformulation-Based Attribution Families  \\
(Unified formulations, representative methods, and key difference)}
\label{tab:Reformulation-based unifications}
\renewcommand\arraystretch{2.5}
\begin{tabular}
{>{\centering} p{23mm} |   >{\raggedright} p{65mm} |  >{\centering} p{56mm} | >{\centering\arraybackslash} p{22mm} }
\toprule
\textbf{Reformu. Family} 
& \makecell[c]{\textbf{Unification Formulation}} 
& \makecell[c]{\textbf{Representative Attribution Methods}}
& \makecell[c]{\textbf{Key Difference}}
\\ \hline

Feature Additive  \cite{lundberg2017unified}
& $\displaystyle a_i = w_i, \ \text{where} \ f(\bm{x}) \approx w_0 + \sum_{i=1}^M w_i z_i \ \textcolor{gray}{(Eq. ~\ref{eqn:additive})} $
& \textit{LRP-$0$/$\epsilon$} \cite{bach2015pixel}, \textit{DL-Res}   \cite{shrikumar2017learning},  
 \textit{Shapley}  \cite{lundberg2017unified},
 \newline
  \textit{LIME}  \cite{ribeiro2016should}
& ---
\\ \hline

Modified GI \cite{ancona2017towards}
& $\displaystyle a_i = \frac{\partial^\mathbb{M} f(\bm{x})}{\partial x_i} \cdot (x_i - b_i) \ \textcolor{gray}{(Eq. ~\ref{eqn:modified})}$
& \textit{LRP-$0$/$\epsilon$}  \cite{bach2015pixel}, \textit{DL-Res}   \cite{shrikumar2017learning}, 
 \textit{Grad$\times$Input} \cite{simonyan2013deep},   \textit{IG} \cite{sundararajan2017axiomatic}
&  Derivative rule ${\partial^{\mathbb{M}} f(\bm{x})}/{\partial x_i}$
\\ \hline

Game-theoretic \quad 
\cite{sundararajan2020many, covert2021explaining, lundstrom2022rigorous}
&
$\displaystyle a_i = \sum\limits_{S \subseteq N \setminus \{i\}}  w_S^{\mathcal{A}}  \cdot [F(\bm{x}_{S \cup \{i\}}) - F(\bm{x}_S)]
 \ (\textcolor{gray}{Eq. ~\ref{eqn:Shapley variant based family})}$
& \textit{IG} \cite{sundararajan2017axiomatic},   \textit{Shapley}  \cite{lundberg2017unified},    \textit{Banzhaf} \cite{dubey1979mathematical}
& set of weights $\{w_S^{\mathcal{A}}\}$
\\ \hline

Path Integral \cite{sundararajan2017axiomatic}
& $ \displaystyle a_i = \int_{t = 0}^1 \frac{\partial f(\gamma(t))}{\partial \gamma_i(t)} \cdot \frac{d \gamma_i(t)}{dt} \, dt
\ (\textcolor{gray}{Eq. ~\ref{eqn:Path integral attribution family})}$
& \textit{IG} \cite{sundararajan2017axiomatic},  
\textit{PathIG} \cite{sundararajan2017axiomatic},  \textit{EG}  \cite{erion2019learning},  \textit{Shapley}  \cite{lundberg2017unified}
& Integration path $\bm{\gamma}(t)$
\\ \hline

Nonnegative BP \cite{sixt2020explanations}
& $\displaystyle \bm{a} = \prod_{l=1}^L M^{(l), +} \bm{y}
\ (\textcolor{gray}{Eq. ~\ref{eqn: Non-negative matrix product})}$
&  \textit{LRP-$\alpha1\beta0$}  \cite{bach2015pixel},   \textit{DTD} \cite{montavon2017explaining}, \textit{ExBP} \cite{zhang2018top},  
\newline
\textit{RectG} \cite{kim2019saliency},  \textit{Deconv} \cite{zeiler2014visualizing}, 
\textit{GBP} \cite{springenberg2014striving}
&  BP matrix  \ \ \ $M^{(l), +}$
\\ \hline

Feature Removal  \cite{covert2021explaining}
&   $\displaystyle \bm{a} = \psi  (\mu(F(\bm{x}_{\emptyset})), \dots, \mu(F(\bm{x}_{N})))
\ (\textcolor{gray}{Eq. ~\ref{eqn:Feature removal attribution family})}$
& 
\textit{Occ-1}  \cite{zeiler2014visualizing},  \textit{Occ-p} \cite{zintgraf2017visualizing},  \textit{PDiff}  \cite{zintgraf2017visualizing},  \textit{Shapley}  \cite{lundberg2017unified}, \textit{Banzhaf} \cite{dubey1979mathematical}, \textit{RISE} \cite{Petsiuk2018rise},  \textit{MP} \cite{fong2017interpretable}, 
\newline
 \textit{EP} \cite{fong2019understanding}, \textit{RTIS} \cite{dabkowski2017real},  \textit{LIME}  \cite{ribeiro2016should}, $\dots$
& Def. of $F(\bm{x}_S)$, \ Behavior $\mu(\cdot)$,  Aggregator $\psi(\cdot)$
\\ \hline

Taylor Interaction \cite{deng2021unified, 10414149}
&  $\displaystyle  a_i =    \sum_{j \in N}  w_{i, j} \cdot  \phi(j) + 
\sum_{S \subseteq N, |S| > 1}  w_{i, S} \cdot I(S) 
(\textcolor{gray}{Eq. ~\ref{eqn:Taylor interaction attribution})}$
& \textit{LRP-$0$/$\epsilon$/$\alpha\beta$} \cite{bach2015pixel}, \textit{DL Res/Rev}  \cite{shrikumar2017learning}, 
\textit{DTD} \cite{montavon2017explaining},  \textit{GradCAM} \cite{selvaraju2017grad}, \textit{IG} \cite{sundararajan2017axiomatic}, \textit{EG} \cite{erion2019learning}, 
 \textit{Grad$\times$Input} \cite{simonyan2013deep},  \textit{Occ-1}  \cite{zeiler2014visualizing},  \textit{Occ-p} \cite{zintgraf2017visualizing},  \textit{Shapley}  \cite{lundberg2017unified}, $\dots$
& Allocation weight $\{w_{i,j}\}, \{w_{i,S}\}$
\\ 
\bottomrule
\end{tabular}
\end{table*}

\subsection{Reformulation-Based Unification}
\label{subsec: Other unification framework}
In contrast to formulation-driven unification, which focuses on the original design principles of attribution methods, reformulation-based unification seeks to reformulate these methods under shared mathematical frameworks.
This sheds light on deeper theoretical connections among attribution methods that are not evident in their standard formulations.

These reformulations give rise to a distinct set of unified attribution families, each grounded in a specific mathematical interpretation, 
such as feature additivity, game-theoretic allocation, or path integrals. 
Table~\ref{tab:Reformulation-based unifications} summarizes the unification formulations, representative methods, and key distinguishing factors of each reformulated family.

\vspace{6pt}
\noindent \textbf{(1) Feature Additive Attribution Family} \cite{lundberg2017unified, covert2020understanding}. 
This family formulates attribution as the coefficients $[w_1, \dots, w_M]$ of a linear surrogate model $g(\bm{z})$ that approximates the local behavior of the DNN $f$ on a given input sample $\bm{x}$:
\begin{equation}\label{eqn:additive}
\begin{aligned}
f(\bm{x}) \approx g(\bm{z}) &= w_0  + \sum\nolimits_{i=1}^M w_i \cdot z_i \  \Rightarrow \ a_i = w_i
\end{aligned}
\end{equation}
Here, each $z_i \in \{0,1\}$ is a binary indicator denoting whether the $i$-th input feature is present or absent in a simplified representation of $\bm{x}$. 
To ensure faithful explanation, $g(\bm{z})$  is required to closely match $f(\bm{x})$.

It has been proven in  \cite{lundberg2017unified, covert2020understanding} that several widely used attribution methods, including \textit{LRP-0/$\epsilon$}, \textit{DeepLIFT Res}, \textit{Shapley value}, and \textit{LIME},
can all be reformulated as the form  in Eq. (\ref{eqn:additive}).
and are thus unified within the feature additive family.

\vspace{6pt}
\noindent \textbf{(2) Modified Gradient$\times$Input Attribution Family} \cite{ancona2017towards}. 
This family computes the attribution $a_i$ as the element-wise product of \textit{modified gradients} and input features:
 \begin{equation}\label{eqn:modified}
\begin{aligned}
a_i = \frac{\partial^{\mathbb{M}} f(\bm{x})}{\partial x_i} \cdot (x_i - b_i)
\end{aligned}
\end{equation}
where $\frac{\partial^{\mathbb{M}} f(\bm{x})}{\partial x_i}$ denotes a modified gradient that replaces the standard derivative rule $\sigma'(z)$ (of non-linear activation functions) in backpropagation with an alternative form, such as $\frac{d^{\mathbb{M}} \sigma(z)}{dz}  = \frac{\sigma(z)}{z}$ or $\frac{ \sigma(z) - \sigma(\bar z) }{z - \bar z}$.
These modifications may enable more accurate attribution particularly in deep networks with complex nonlinearities. 

It has been proven in \cite{ancona2017towards} that  \textit{LRP-$0$/$\epsilon$}, \textit{DeepLIFT Res}, \textit{Grad$\times$Input}, and \textit{IG} can all be reformulated into the unified form in Eq.~(\ref{eqn:modified}).
The \textbf{main difference} among them lies in the specific definition of modified gradient 
 ${\partial^{\mathbb{M}} f(\bm{x})}/{\partial x_i}$.

In particular, several equivalences among certain attribution methods have been established:
\begin{itemize}
  \item \textit{LRP-$\epsilon$} is equivalent to \textit{Grad$\times$Input} if and only if  ReLU is used as the activation function.
  \item \textit{LRP-$\epsilon$} and \textit{DeepLIFT Res} are equivalent under zero-bias networks with homogeneous nonlinearities satisfying $\sigma(0)=0$ (e.g., ReLU or Tanh).
\end{itemize}
These reformulations have led to more efficient implementations of these methods\footnote{\url{https://github.com/marcoancona/DeepExplain}}.

\vspace{6pt}
\noindent \textbf{(3) Game-Theoretic Attribution Family}  \cite{sundararajan2020many, covert2021explaining, lundstrom2022rigorous}.    
This family is rooted in cooperative game theory, which feature attributions are derived by allocating marginal contributions across all feature subsets.  
Attribution for a feature $i$ is defined as a weighted average of its marginal contributions over all coalitions $S \subseteq N \setminus \{i\}$:
\begin{equation}\label{eqn:Shapley variant based family}
a_i = \sum\limits_{S \subseteq N \setminus \{i\}} w_S^{\mathcal{A}} \cdot \left[F(\bm{x}_{S \cup \{i\}}) - F(\bm{x}_S)\right],
\end{equation}
where  the weights $w_S^{\mathcal{A}}$ are uniquely determined by a specified set of axioms $\mathcal{A}$ (e.g., linearity, dummy, symmetry, efficiency).

Different instantiations of $\mathcal{A}$ yield distinct game-theoretic attribution methods. 
\begin{itemize}
  \item \textit{Shapley Value} \cite{lundberg2017unified}, the unique method satisfying linearity, dummy, symmetry, and efficiency axioms over discrete function spaces.
  
  \item \textit{Integrated Gradients (IG)} \cite{sundararajan2017axiomatic}, equivalent to the Aumann–Shapley value, the unique method additionally satisfying implementation invariance axiom over differentiable function spaces.
  
  \item \textit{Banzhaf Value} \cite{dubey1979mathematical}, the unique method satisfying linearity, dummy, symmetry, and 2-efficiency axioms.
\end{itemize}
The \textbf{key distinction} among these methods lies in the choice of axioms, which determines both theoretical guarantees and practical behavior.

\vspace{6pt}
\noindent \textbf{(4) Path Integral Attribution Family}. 
This family defines the attribution $a_i$ as the integral  of the gradients along a path $\gamma$ from the baseline to the input  \cite{sundararajan2017axiomatic}:
\begin{equation}\label{eqn:Path integral attribution family}
\begin{aligned}
a_i &= \int_{t = 0}^1 \frac{\partial f(\gamma(t))}{\partial \gamma_i(t)} \cdot \frac{d \gamma_i(t)}{d t} dt 
\end{aligned}
\end{equation}
\noindent where $\gamma(t) = [\gamma_1(t), \dots, \gamma_n(t)]: [0,1] \rightarrow \mathbb{R}^n$ is a path from the baseline $\bm{b}$ to the input $\bm{x}$, such that $\gamma(0) = \bm{b}$ and $\gamma(1) = \bm{x}$. 
This formulation captures the intuition of continuously accumulating feature contributions along its interpolation trajectory.

Canonical methods in this family include \textit{Integrated Gradients (IG)}, \textit{PathIG}, and \textit{Expected Gradients (EG)}.
Furthermore, as noted in \cite{sundararajan2017axiomatic}, \textit{Shapley value} can be viewed as discrete counterparts of path integral methods: while \textit{IG} integrates gradients along a continuous path, \textit{Shapley value} instead accumulate marginal contributions over numerous collections of discrete paths.
In addition, more path integral variants have been proposed in recent work \cite{yang2023local, jeon2023beyond, zhuo2024ig}.
The \textbf{main difference} among these methods lies in the specification of the integration path $\gamma$.  
\\

\vspace{6pt}
\noindent \textbf{(5) Nonnegative Backpropagation Attribution Family} \cite{sixt2020explanations}.  
This family encompasses layer-wise BP attribution methods that propagate relevance signals backward using nonnegative transformations. 
Based on Eq.~(\ref{eqn:layer-wise backpropagation based attribution family}), such propagation can be generally expressed as a product of non-negative backpropagation matrices across different layers:
\begin{equation}\label{eqn: Non-negative matrix product}
\bm{a} = \prod\nolimits_{l=1}^L M^{(l), +} \bm{y}
\end{equation}
\noindent where the nonnegative matrix $M^{(l), +}$ satisfies $M_{i,j}^{(l), +} \geq 0$ for all entries. This property ensures that if $a_j^{(l)}$ is a positive (or negative) attribution, it propagates only its positive (or negative) components to attributions $a_i^{(l-1)}$ in preceding layers.

It has been demonstrated in \cite{sixt2020explanations} that several popular dmethods—including \textit{LRP-$\alpha1\beta0$}, \textit{DTD}, \textit{ExBP}, \textit{RectG}, \textit{Deconv}, and \textit{GBP}—can all be unified under this formulation. The \textbf{main difference} among them lies in the construction of the nonnegative backpropagation matrix $M^{(l), +}$, which encodes specific propagation rules.

\vspace{6pt}
\noindent \textbf{(6) Feature Removal Attribution Family} \cite{covert2021explaining}.  
This family estimates feature importance by removing (i.e., masking) subsets of input features and observing the model's output changes. 
Each method is unified by the following general formulation:
\begin{equation}\label{eqn:Feature removal attribution family}
\begin{aligned}
\bm{a} = \psi & (\mu(F(\bm{x}_{\emptyset})), \dots, \mu(F(\bm{x}_{N}))), \\
\text{where } \ F(\bm{x}_S) & = \mathbb{E}_{\bm{b}_{\bar S} \sim p(\bm{b}_{\bar S})}[f(\bm{x}_S)]
\end{aligned}
\end{equation}
\noindent where $\bm{x}_S$ denotes the masked sample where features in $\bar{S}$ have been replaced by baseline values $\bm{b}_{\bar{S}}$.
The function $F(\bm{x}_S)$ represents the expected network output $f(\bm{x}_S)$ over different baselines, sampled from a predefined distribution $p(\bm{b}_{\bar{S}})$.

This formulation contains three key components:
\begin{itemize}
\item 
\textbf{Baseline distribution $p(\bm{b}_{\bar S})$:} defines how to select baseline values for replacement when removing features. Typical settings include:
\begin{equation}\label{eqn:baseline distribution}
\!\! \!\! 
p(\bm{b}_{\bar S})  =  \left
\{
\begin{array}{ll}
\delta(\bm{b}_{\bar S}) & \text{(fixed baseline)} \\
p(\bm{x}_{\bar S}) & \text{(marginal distribution)} \\
p(\bm{x}_{\bar S}|\bm{x}_S) & \text{(conditional distribution)} \\
\end{array}
\right.
\end{equation}
where $\delta(\bm{b}_{\bar S})$ denotes a Dirac delta distribution centered at a fixed baseline.

\item \textbf{Behavior function $\mu(\cdot)$:} specifies the model behavior of interest, such as prediction probability $\mu = F(\bm{x}_S)$ or negative loss $\mu = - \mathcal{L}(F(\bm{x}_S), y)$ \cite{lundberg2020local, schwab2019cxplain}.

\item \textbf{Aggregator $\psi(\cdot)$:}  determines how to summarize contributions across subsets $S$, e.g., leave-one-out (e.g., \textit{Occ-1}), Shapley average (e.g., \textit{Shapley value}, \textit{SAGE}), Banzhaf average (e.g., \textit{Banzhaf value}), mean when included (e.g., \textit{RISE}),  subset optimization (e.g., \textit{MP}, \textit{EP}).
\end{itemize}

It has been shown in \cite{covert2021explaining} that 26 existing attribution methods, widely used ones such as \textit{Occ-1}, \textit{Occ-p}, \textit{PDiff}, \textit{Shapley Value}, \textit{SAGE}, \textit{Banzhaf Value}, \textit{RISE}, \textit{MP}, \textit{EP}, \textit{RTIS}, \textit{LIME}, \textit{L2X} \cite{strumbelj2010efficient}, \textit{LossSHAP} \cite{lundberg2020local}, and \textit{CXPlain} \cite{schwab2019cxplain}, can be expressed in this unified form in Eq.~(\ref{eqn:Feature removal attribution family}).
The \textbf{main differences} among them arise from the three elements above.

\begin{figure*}[t]\centering
\includegraphics[width = 0.92 \textwidth]{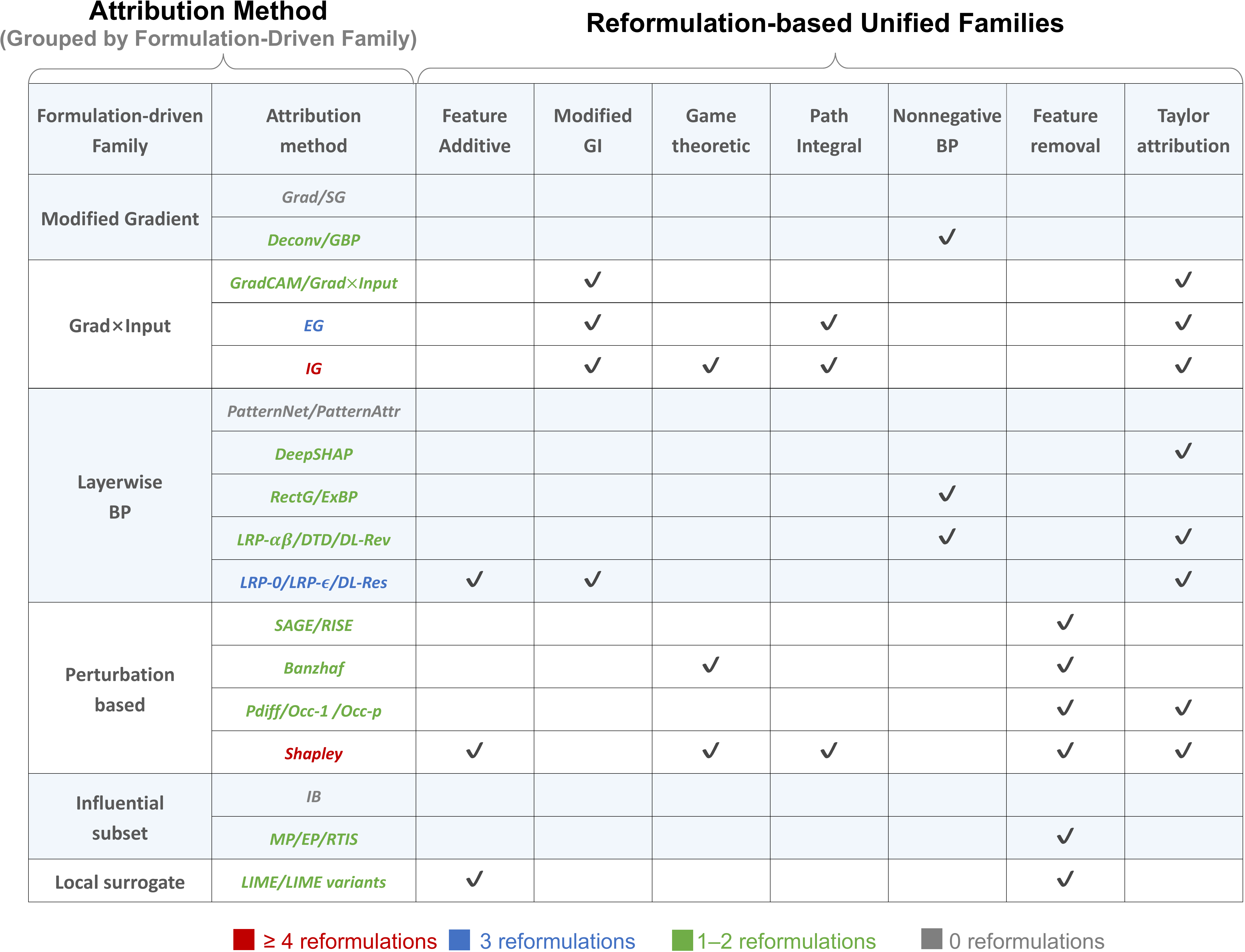}
\vspace{-5pt}
\caption{
Structural mapping between formulation-driven attribution families and reformulation-based unified families.
Each row represents an attribution method grouped by its original (formulation-driven) family, 
while checkmarks indicate its inclusion in various reformulation-based families (columns).
The color of each method name denotes the number of reformulations it participates in: red ($\geq 4$), blue ($3$), green ($1-2$), and gray ($0$).
This mapping highlights how certain methods, such as \textit{IG} and \textit{Shapley}, serve as central connectors across the theoretical landscape.
It is worth noting that this mapping may not capture all associations—some methods could conceptually belong to certain reformulation families but remain explicitly uncategorized in prior works due to scope limitations or overlooked theoretical links.
}
\label{fig: unification_method_map}
\end{figure*}

\vspace{6pt}
\noindent \textbf{(7) Taylor Interaction Attribution Family} \cite{10414149, deng2021unified}.  
This family derives from the multivariate Taylor expansion of the model output around a baseline $\bm{b}$.  
Based on Taylor theorem, Deng et al. \cite{deng2021unified} formally proved that the network output $f(\bm{x})$ can be decomposed into two disjoint components:
(i) \textit{independent effects} $\phi(j)$, quantifying the individual contributions of each input variable $x_j$; 
and (ii) \textit{interaction effects} $I(S)$, capturing the interaction effects resulting from the collaboration among multiple input variables in the subset $S$.
This yields a complete decomposition of the model output:
\begin{equation}\nonumber 
\begin{aligned}
f(\bm{x}) = f(\bm{b}) + \sum_{j \in N} \phi(j) + \sum_{S \subseteq N, |S|>1} I(S)
\end{aligned}
\end{equation}

Building on this decomposition, this family  formulates $a_i$ as a weighted aggregation of both types of effects:
\begin{equation}\label{eqn:Taylor interaction attribution}
\begin{aligned}
 a_i =    \sum_{j \in N}  w_{i, j} \cdot  \phi(j) + 
\sum_{S \subseteq N, |S| > 1}  w_{i, S} \cdot I(S) 
\end{aligned}
\end{equation}
\noindent where $w_{i, j}, w_{i,S}$ denote the allocation weights of independent and interaction effects to input $x_i$, respectively.

It has been proven in \cite{10414149} that 14 popular attribution methods, including \textit{Grad$\times$Input},  \textit{IG}, \textit{EG}, \textit{GradCAM},  \textit{LRP-$\epsilon$}, \textit{LRP-$\alpha\beta$},  \textit{DeepLIFT Res}, \textit{DeepLIFT Rev}, \textit{DTD}, \textit{Occ-1},  \textit{Occ-p},  \textit{PDiff},   \textit{Shapley value}, \textit{Deep SHAP}, can all be unified within this Taylor interaction attribution family.
The \textbf{main difference} among them lies in how the weighting scheme $\{w_{i, j}\}, \{w_{i, S}\}$ are defined to allocate independent and interaction effects.

\subsection{Our Insights: Lessons from Theoretical Unification}  
\label{subsec:our perspective1}
Theoretical unification offers a principled lens to revisit attribution methods. Below, we summarize three core insights derived from our dual-perspective unification analysis.
\vspace{2pt}

\noindent \textbf{(1) Facilitating in-depth understanding of attribution methods.} 
Theoretical unification enhances our understanding of attribution methods in three key aspects:

(i) \textit{Unified understanding of diverse methods}.
While attribution methods often appear diverse—built upon distinct heuristic designs—theoretical unification frameworks uncover their underlying commonalities.
For example, all 14 methods under the Taylor interaction family, despite different motivations, can be reformulated as weighted sums of independent and interaction effects.  
This reveals that their essential differences lie not in implementation details, but in how they assign these effects to input variables.

(ii) \textit{Multi-perspective understanding of attribution methods.}
Each unification framework provides a unique  theoretical lens for interpreting attribution methods. 
Taken together, these frameworks enable a multi-perspective understanding, revealing how a single method can embody diverse explanatory intuitions and thus contribute to a more holistic  understanding of the attribution landscape.
As shown in Figure~\ref{fig: unification_method_map}, the \textit{Shapley value} aligns with five reformulation families—feature additive, game-theoretic, path integral, feature removal, and Taylor interaction—each highlighting different aspects, ranging from additive contribution semantics to effect allocation logic.

(iii) \textit{Principled subtyping via structural alignment}.
While formulation-driven grouping provides a coarse taxonomy, reformulation-based unification  allows for  a finer subtyping based on shared theoretical structure.  
Specifically, within the same formulation-driven family, methods that share identical reformulation  families can be identified as a principled subtype, exhibiting stronger structural and conceptual cohesion.

As shown in Figure~\ref{fig: unification_method_map},  within layerwise BP family, methods such as \textit{LRP-$\alpha\beta$}, \textit{DTD}, and \textit{DL-Rev} are all jointly unified under Nonnegative BP and Taylor interaction attribution families.  
This reflects a shared polarity-aware design that explicitly separate positive and negative contributions during backpropagation,  distinguishing them from other layerwise methods lacking this structural refinement.
\vspace{4pt}

\noindent \textbf{(2) Enabling theoretically grounded comparison and evaluation.}  
Beyond deepening theoretical understanding, unification facilitates systematic and scalable theoretical evaluation.  
Without shared formulations, theoretical evaluation often requires isolated, method-specific proofs—a labor-intensive and fragmented process.  
Unification frameworks instead group methods under common structures, enabling evaluation at the attribution family level.  
This shift supports generalized analysis for faithfulness, robustness, and other comparative behaviors, enhancing both rigor and efficiency.
\vspace{4pt}

\noindent \textbf{(3) Serving as an auxiliary metric for evaluating theoretical soundness.}  
Unification itself can act as an auxiliary metric for theoretical evaluation—revealing how many theoretical perspectives a method is compatible with.  
Such compatibility provides a signal of theoretical soundness, as methods align with more reformulation families are more likely to capture fundamental principles shared across distinct paradigms.  
For example, \textit{Shapley value} aligns with five out of seven reformulation families, and \textit{IG} aligns with four, suggesting their strong theoretical generality.
However, it is worth noting that while broad compatibility implies stronger theoretical soundness, it does not guarantee practical faithfulness.

\begin{figure*}[t]\centering
\includegraphics[width = 0.95 \textwidth]{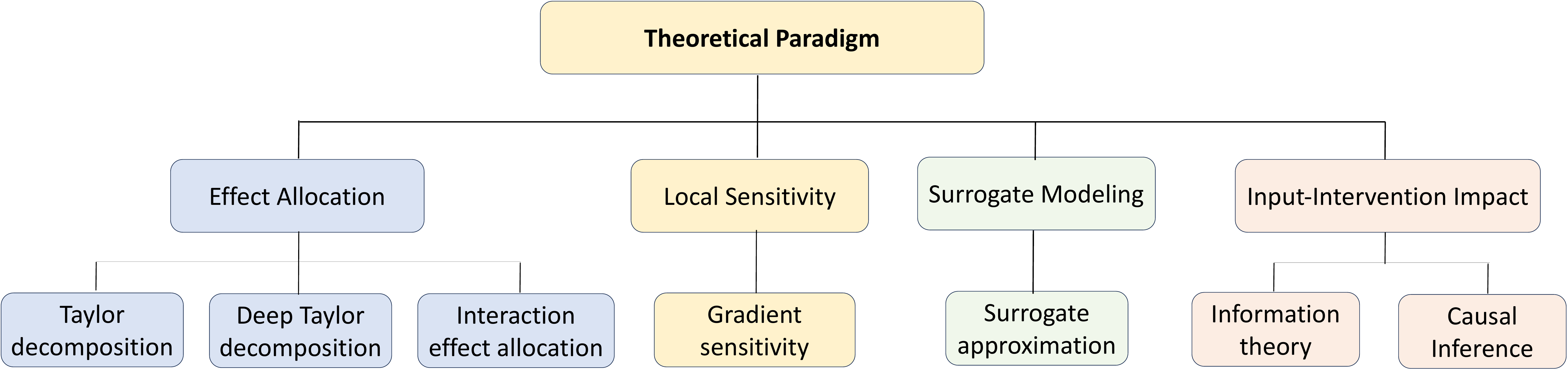}
\caption{Conceptual map summarizing prior theoretical rationales behind attribution methods.}
\label{fig:rationale overview}
\end{figure*}

\section{Theoretical rationale}
\label{Sec:theoretical rationale}
Although numerous attribution methods have been proposed, most of them are heuristic, with their theoretical foundations either unspecified or unverified. 
Recent efforts have introduced diverse theoretical rationales to explain the mechanisms underlying attribution methods, i.e., \textit{why a given method provides a meaningful estimation of feature importance}, 
thus offering a more principled view of attribution.

To provide a structured review, we organize these rationales into four complementary paradigms, each offering a distinct perspective on how input variables contribute to outputs:
\begin{itemize}
    \item  \textbf{Local Sensitivity (Section \ref{subsec:Intrinsic sensitivity})}, attributes importance based on how sensitively the model output responds to local input perturbations;

\item  \textbf{Effect Allocation (Section \ref{subsec:Effect Allocation})}, infers attribution by decomposing model output into additive, identifiable effects and assigning them to input variables;
    
\item  \textbf{Surrogate Modeling (Section \ref{subsec:Surrogate Modeling})},  employs locally interpretable models to approximate attribution;

\item  \textbf{Input-Intervention Impact (Section \ref{subsec:Input-Intervention Impact})}, quantifies importance through interventions on inputs, often grounded in causal or information-theoretic principles.
\end{itemize}

Beyond summarizing existing works, Section~\ref{subsec:our perspective2} presents our insights into how these rationales provide principled justifications for attribution faithfulness.

\subsection{Local sensitivity}
\label{subsec:Intrinsic sensitivity} 
The \textit{Local Sensitivity} paradigm attributes feature importance by quantifying how sensitively the model output responds to small perturbations at specific input points.

\vspace{6pt}
\noindent\textbf{(1) Gradient Sensitivity}.  
This rationale measures importance based on the model’s local sensitivity to infinitesimal input perturbations. Specifically, the gradient $\partial f(\bm{x})/\partial x_i$ quantifies how much the model output $f(\bm{x})$ changes in response to infinitesimal changes in the input feature $x_i$.
A larger gradient implies  a stronger local influence of that feature on the model’s prediction \cite{baehrens2010explain, srinivas2021rethinking, shah2021input}. Representative methods such as \textit{Gradient} \cite{baehrens2010explain} directly adopt this rationale by using the gradient vector as the attribution score.

This rationale is widely regarded as the foundational justification for many methods in the \textit{Modified Gradient family}, where gradients are often adjusted or augmented to enhance attribution quality and stability.

\subsection{Effect Allocation}
\label{subsec:Effect Allocation}
The \textit{Effect Allocation} paradigm assigns importance by formally decomposing the model output into additive, identifiable effects and allocating/attributing them to input variables according to different principles.
This paradigm encompasses three distinct theoretical rationales, each offering a different decomposition logic and allocation mechanism.

\noindent (i) \textit{Taylor decomposition} attributes importance by linearly decomposing the output changes into feature-wise attributions via first-order Taylor expansion.

\noindent (ii) \textit{Deep Taylor decomposition} recursively propagates relevance through the network by applying localized first-order Taylor expansions at each layer.

\noindent (iii) \textit{Taylor interaction allocation} distributes both independent and interaction effects to input features based on a structured higher-order Taylor expansion framework.

\vspace{8pt}
\noindent\textbf{(1) Taylor Decomposition}.  
The Taylor decomposition rationale interprets attributions by locally linearizing the function via first-order Taylor expansion and attributing the output change to individual input variables based on the resulting linear expansion terms.

\textit{Grad$\times$Input} \cite{simonyan2013deep} is a representative method that instantiates this rationale.
Specifically, \textit{Grad$\times$Input} can be interpreted as performing a first-order Taylor expansion at the input $\bm{x}$ with respect to a baseline $\bm{b} = \bm{0}$ \cite{montavon2017explaining}:
\begin{equation}\label{eqn:Taylor decomposition}
f(\bm{b}) = f(\bm{x}) + \sum\nolimits_{i} \frac{\partial f(\bm{x})}{\partial x_i} \cdot (b_i - x_i) + \epsilon_1,
\end{equation}
and then allocating the corresponding decomposed effect $a_i = {\partial f(\bm{x})}/{\partial x_i} \cdot (x_i - b_i)$ to each input feature $x_i$.

This rationale underlies many attribution methods within the \textit{Gradient$\times$Input attribution family} \cite{samek2021explaining}, such as  \textit{IG}, offering a unified explanation for how these methods decompose and allocate model outputs.

\begin{table*}[t]
\centering
\caption{Theoretical rationales of attribution methods: summary, representative methods, and known limitations.}
\label{tab:theoretical_rationales}
\renewcommand{\arraystretch}{2.5}
\begin{tabular}{>{\centering}p{33mm} | >{\centering} p{68mm} | >{\centering}p{30mm} | 
 >{\centering\arraybackslash}  p{32mm}}
\toprule
\textbf{Theoretical Rationale}
& \textbf{Rationale Summary}
& \textbf{Representative Methods}
& \textbf{Associated Family} \\
\hline

Gradient Sensitivity \quad \quad 
\cite{baehrens2010explain,srinivas2021rethinking,shah2021input} 
 &
Measures the sensitivity of the network output responds to small input perturbations
&
\textit{Gradient} &
Modified Gradient family \\

\hline
Taylor Decomposition  \quad \cite{samek2021explaining,montavon2017explaining} &
Linearly decomposing the output changes into feature-wise attributions via first-order Taylor expansion
&
\textit{Grad$\times$Input}, \textit{IG} &
Gradient×Input family \\

\hline
Deep Taylor Decomposition \cite{samek2021explaining, montavon2019layer}  &
Recursively conducts Taylor decomposition in a layer-wise manner to compute attributions
&
\textit{DTD}, \textit{LRP-0/$\epsilon$} &
Layerwise BP family
\\
\hline

Interactions  Effect Allocation \cite{10414149,deng2021unified} &
Decomposes output changes into independent and interaction effects, and 
distributing them to input variables
&
\textit{All Taylor interaction methods} &
Taylor Interaction family
\\
\hline

Surrogate Approximation \cite{ribeiro2016should,garreau2020explaining} &
Fits a simple surrogate model (e.g., linear regressor) to approximate DNN's local behavior  
& \textit{LIME}, \textit{Shapley} &
Local Surrogate family
 \\
\hline

Causal Attribution \cite{chattopadhyay2019neural} &
Analyzes how explicit interventions on input variables directly lead to measurable changes in the model output
& \textit{ACE, Shapley, Occ-1} &
Perturbation-based  family
\\ 
\hline

Information Theory  \quad \quad \cite{covert2021explaining, schulz2020restricting,deng2021mutual}
&
Evaluates how much predictive information a specific feature or feature subset contributes to the final output
& \textit{SAGE, IB} 
& Feature Removal family
\\
\hline
\end{tabular}
\end{table*}

\vspace{8pt}
\noindent\textbf{(2) Deep Taylor Decomposition}.
\label{subsec:Layer-wise Taylor decomposition}
This rationale infer attributions by recursively applying first-order Taylor expansions at each layer of the network, thereby propagating output relevance back to input features in a layer-wise fashion.
By localizing the decomposition to individual layers, this approach reduces the approximation errors commonly associated with global Taylor expansions over the entire network.

Representative methods such as \textit{DTD} \cite{montavon2017explaining}, instantiate this rationale as follows.
For a DNN $f(\bm{x}) = f^{(L)}(\cdots f^{(1)}(\bm{x}))$, each layer performs a Taylor expansion of a neuron $x_j^{(l)}$ around a baseline $\bm{b}^{(l-1)}$ (omit layer subscripts for brevity):
\begin{equation}
f_j(\bm{x}) = f_j(\bm{b}) + \sum_i \frac{\partial f_j(\bm{b})}{\partial x_i} (x_i - b_i) + \epsilon_1.
\end{equation}
The relevance $a^{(l)}_{i \rightarrow j}$ from $x_i^{(l-1)}$ to $x_j^{(l)}$ is defined as the linearized term, i.e., ${\partial f_j(\bm{b})}/{\partial x_i} \cdot (x_i - b_i)$, 
and attributions are propagated by summing over all  connected neurons.
The \textit{DTD} method further specifies baseline selection rules (e.g., $w^2$-rule, $z^+$-rule) designed to minimize the expansion error $\epsilon_1$ and improve attribution faithfulness \cite{samek2021explaining, montavon2017explaining}.

This rationale broadly underpins attribution methods in the \textit{Layerwise BP attribution family}, even when not explicitly analyzed.
Notably, \textit{LRP-0} and \textit{LRP-$\epsilon$} can be viewed as special cases of \textit{DTD}, corresponding to particular baseline selections, and are thus grounded in this rationale \cite{montavon2019layer}.

\vspace{8pt}
\noindent\textbf{(3) Interaction Effect Allocation.}  
This rationale attributes importance by decomposing the model output into both the independent effects of individual input variables and the interaction effects arising from joint subsets of variables, and then distributing these effects to the corresponding input features.  
Unlike first-order methods such as Taylor decomposition and Deep Taylor decomposition, which primarily capture independent effects via local linear approximations, this rationale accounts for both higher-order individual contributions and complex interaction effects, offering a more comprehensive explanation of feature attributions.

It has been shown in \cite{10414149, deng2021unified} that all methods within the \textit{Taylor interaction attribution family} adhere to this rationale by explicitly or implicitly allocating both independent and interaction effects.
As a concrete example, consider the \textit{Occ-1} method and the \textit{Shapley value} method.  
The \textit{Occ-1} method assigns the full contribution of all interaction effects involving input variable $i$ to $i$ itself,  
whereas the \textit{Shapley value} method equally distributes the contribution of each interaction term among all variables in the subset $S$.
\begin{equation}
\begin{aligned}
 a_i^{\text{Occ-1}} &  = \phi(i) + \sum\nolimits_{S\subseteq N, |S|>1, S \ni i}I(S)\\
a_i^{\text{Shapley}} & =  \phi(i) + \sum\nolimits_{S\subseteq N, |S|>1, S \ni i} \frac{1}{|S|} I(S)
\end{aligned}
\end{equation}
where $S \ni i$ denotes that the interaction term $I(S)$ involves input variable $i$.

\subsection{Surrogate Modeling}
\label{subsec:Surrogate Modeling}
In this subsection, we introduce the \textit{Surrogate Modeling} paradigm, which computes feature attributions based on local approximation.  
This paradigm assumes that the complex DNN can be locally approximated by a simpler, interpretable surrogate model (e.g., a linear regressor).  
Feature importance is then inferred from the parameters or structure of the surrogate model.
\\

\noindent\textbf{(1) Surrogate Approximation}.
\label{subsec:Local surrogate}
This rationale explains the prediction of a DNN $f(\bm{x})$ by approximating it locally  with a simpler surrogate model $g(\bm{x})$, thus transforming the attribution task into a more tractable problem:
\begin{equation}\label{eqn:surrogate model}
g(\bm{x}) \approx f(\bm{x}) \quad \Rightarrow \quad \bm{a}^g \approx \bm{a}^f.
\end{equation}

A representative example is the \textit{LIME} method \cite{ribeiro2016should}, which fits a simple linear surrogate model $g(\bm{x}) = \bm{w}^T \bm{x}$ within a local neighborhood $\mathcal{N}_{\bm{x}}$ around input. The learned coefficients $\bm{w}^*$ then serve as attribution scores. This rationale forms the theoretical foundation for the \textit{local surrogate} attribution family. In addition, several works further reinforce the soundness of \textit{LIME} from theoretical and statistical validity  perspectives \cite{garreau2021does, garreau2020explaining, garreau2020looking, mardaoui2021analysis}.

Beyond the local surrogate family, this rationale is also implicitly reflected in other attribution methods.
Notably, the \textit{Shapley value} can be viewed as a special case of surrogate approximation.
As shown in \cite{lundberg2017unified}, it also fits a linear surrogate model $g(\bm{x}) = \bm{w}^T \bm{x}$, but estimates coefficients by averaging over all possible feature subsets $\bm{x}_S$ (for all $S \subseteq N$).

\subsection{Input-Intervention Impact}
\label{subsec:Input-Intervention Impact}
The \textit{Input-Intervention Impact} paradigm quantifies feature importance by evaluating how explicit interventions on input variables affect the model's predictions. 
This  paradigm measures how interventions to input features (e.g., perturbations) change the output, and attributes the resulting change to the corresponding input variables.
This paradigm encompasses two distinct rationales:

\noindent (1) \textit{Causal inference} quantifies the causal effect of each input variable  by analyzing how explicit interventions on input variables directly result in measurable changes in model output. This rationale seeks to identify genuine causal contributions of input features on outputs.

\noindent (2) \textit{Information theory} measures feature importance by evaluating how much predictive information a specific feature or feature subset contributes to the final prediction. Common metrics include mutual information, entropy reduction, and conditional entropy, which quantify how the presence or absence of features affects the uncertainty in predictions.
\\

\noindent \textbf{(1) Causal Inference}. 
Causal inference has become a prominent branch of research in explainable AI \cite{carloni2025role}. In the context of attribution, it provides a more principled rationale for understanding attribution methods. Specifically, it goes beyond simple statistical association by evaluating the causal necessity of input features—i.e., determining whether intervening on an input variable leads to a change in the model's prediction.

A representative method for this rationale is \textit{Average Causal Effect (ACE)} \cite{chattopadhyay2019neural}, which estimates the expected change in the model output when an input feature $x_i$ is intervened upon, by marginalizing over all other variables. This is done through controlled input perturbations, as shown below:
\begin{equation}
\begin{aligned}
 a_i &  = \mathbb{E}[f(\bm{x}) \mid do(x_i = \alpha)] - \mathbb{E}[f(\bm{x}) \mid do(x_i = b_i)] \\
& = \int \left[f(\bm{x} \mid x_i =
\alpha) - f(\bm{x} \mid x_i = b_i)\right] \cdot p(x_{\setminus i}) dx_{\setminus i}
\end{aligned}
\end{equation}
Moreover, the \textit{Shapley value} can be interpreted as a generalized and robust extension of the \textit{ACE} framework. While both aim to quantify the expected impact of an input feature under interventions, the \textit{Shapley value} does so by averaging marginal contributions across all possible contextual feature subsets. 
This subset-based aggregation makes it a more comprehensive causal measure. 
Furthermore, Watson et al. \cite{watson2021local} demonstrate that \textit{Shapley value} closely aligns with the \textit{Probability of Sufficiency (PS)} in causal theory, which denotes the probability that a feature alone would be sufficient to produce the outcome.

In contrast to \textit{ACE}, most perturbation-based methods such as \textit{Occ-1} implicitly compute the \textit{Individual Causal Effect (ICE)}, which evaluates the change in the model's output for a specific input $\bm{x}$ without marginalizing over other input variables: 
$ICE_i(\bm{x}) = f(\bm{x} | do(x_i = \alpha) - f(\bm{x} | do(x_i = b_i)$. 
While \textit{ICE} is useful for localized, instance-specific causal analysis, it does not account for feature interactions or the broader context of input variables. As a result, it may fail to capture the complete causal influence of the intervened feature.
\\

\noindent\textbf{(2) Information Theory}.
Information theory provides a theoretical foundation for attributing importance based on how much information an input feature conveys about the model’s prediction. 
This rationale underpins the \textit{feature removal attribution family}, where attribution is computed by intervening on input variables (typically via removal) and measuring the resulting reduction in predictive information.

Recent studies \cite{covert2021explaining} demonstrate that all methods within the feature removal family  conform to this rationale, provided features are properly removed. 
In particular, the behavior function $\mu(\cdot)$ of each method determines how outputs from partial inputs $\bm{x}_S$ are interpreted in information theory.  
For example, when $\mu(F(\bm{x}_S)) = F(\bm{x}_S)$ uses  network predictions, the method is associated with the conditional probability $p(y \mid X_S = \bm{x}_S)$—i.e., the likelihood of the target $y$ given partial observation $\bm{x}_S$.  
Alternatively, when $\mu(F(\bm{x}_S)) = -\mathcal{L}(F(\bm{x}_S), y)$ adopts the negative loss, the attribution aligns with point-wise mutual information $MI(y, \bm{x}_S)$, which reflects the uncertainty reduction about $y$ after observing $\bm{x}_S$:
\begin{equation}
\begin{aligned}
\text{if } \mu(F(\bm{x}_S)) &= F(\bm{x}_S),   \ \ \bm{a} \rightarrow p(y \mid X_S = \bm{x}_S) \\
\text{if }  \mu(F(\bm{x}_S)) &= -\mathcal{L}(F(\bm{x}_S), y), \  \ \bm{a}  \rightarrow MI(y, \bm{x}_S)
\end{aligned}
\end{equation}

Some feature removal methods go beyond the above formulations and support richer information-theoretic interpretations.  
For instance, \textit{SAGE} estimates attributions  as the weighted average of conditional mutual information $MI(Y, x_i \mid \bm{x}_S)$, quantifying the expected reduction in uncertainty when $x_i$ is added to a given subset $\bm{x}_S$ \cite{covert2020understanding}.  
Additionally, methods such as \textit{MP} and \textit{IB} formulate attribution as a feature subset selection problem, seeking subsets $S$ that maximize mutual information with the output \cite{schulz2020restricting, deng2021mutual}, i.e., $S^* = \mathop{\arg\max}_S \ MI(\bm{x}_S, y) + \lambda \cdot R(S)$.  
Here, the regularization term $R(S)$ controls subset sparsity or redundancy.

\begin{table}[t]
\centering
\caption{Concerns or limitaitons of certain theoretical rationales.}
\label{tab:concerns or limitations}
\renewcommand{\arraystretch}{2}
\begin{tabular}{>{\centering}p{33 mm} | p{49 mm}}
\toprule
\textbf{Theoretical Rationale} 
& \makecell[c]{\textbf{Concerns or Limitations}} \\
\hline

Gradient Sensitivity 
 &
Gradients neglect global importance due to model saturation phenomenon \cite{smilkov2017smoothgrad,shrikumar2017learning} \\

\hline
Taylor Decomposition 
&
First-order expansion error may be non-negligible in complex DNNs \cite{nie2018theoretical,deng2021unified} \\

\hline
Deep Taylor Decomposition 
&
Can reduce to Taylor decomposition or yield produce arbitrary attributions
\cite{sixt2022rigorous}
\\
\hline

Surrogate Approximation   &
Approximation error \cite{tan2024glime, garreau2020explaining}; 
instability problem \cite{bansal2020sam,zhao2021baylime,visani2020optilime,zhou2021s} 
 \\
\hline
\end{tabular}
\end{table}

\subsection{Our Insight: Lessons from Theoretical Rationales}
\label{subsec:our perspective2}
In this subsection, we present key insights drawn from our analysis of theoretical rationales.  
In particular, we identify two critical dimensions for evaluating the theoretical utility of attribution rationales:
(i) the intrinsic soundness of a rationale; 
and
(ii) the fidelity with which the rationale is instantiated in actual attribution algorithms.
\vspace{4pt}

\noindent \textbf{(1) Soundness of Attribution Rationales}.  
The actual soundness of rationales depends on the strength and generality of their underlying assumptions.  
Several rationales exhibit intrinsic weaknesses that may limit  their reliability in practice, which is summarized in Table \ref{tab:concerns or limitations}:
\begin{itemize}
  \item \textit{Local sensitivity}-based rationales neglect global feature importance. 
  Due to the model saturation phenomenon in DNNs \cite{smilkov2017smoothgrad, shrikumar2017learning}, features with small gradients may still exert significant influence on predictions—rendering local gradient-based explanations insufficient.
  
  \item \textit{Taylor decomposition} is vulnerable to large first-order approximation errors ($\epsilon_1$), particularly in highly non-linear models like DNNs \cite{nie2018theoretical, deng2021unified}. This undermines its ability to accurately capture feature contributions.

  \item The theoretical soundness of \textit{Deep Taylor Decomposition}  rationale is under debate. A recent theoretical analysis \cite{sixt2022rigorous} reveals that:  
  (i) when using constant baselines $\bm{b}$, DTD collapses to \textit{Grad$\times$Input}, offering no additional benefit compared to standard Taylor decomposition;  
  (ii) when using  input-dependent baselines, DTD can be manipulated to produce arbitrary attributions, raising concerns about consistency and theoretical soundness.

  \item For \textit{surrogate modeling}, attribution reliability depends on the surrogate model’s approximation fidelity \cite{tan2024glime}. Studies show that \textit{LIME} often incurs non-negligible approximation error when applied to DNNs on tabular data \cite{garreau2020explaining}. Moreover, its reliance on randomly sampled neighborhoods introduces instability  \cite{zhao2021baylime, visani2020optilime, zhou2021s, bansal2020sam}.
\end{itemize}

\noindent In contrast, the \textit{interaction effect allocation}, \textit{causal inference}, and \textit{information theory} rationales are grounded in more general probabilistic or game-theoretic principles and rely on fewer model-specific assumptions. As a result, they benefit from more mature theoretical foundations and tend to exhibit greater soundness across diverse attribution scenarios.
\\

\noindent 
\textbf{(2) Rationale-Fidelity of Attribution Methods}.  
Even when a rationale is theoretically sound, it remains critical to assess how faithfully it is instantiated in specific attribution methods.  
Some methods loosely adopt the core idea of a rationale without strict adherence to its formalism, which may compromise reliability.  
Others, by contrast, closely adhere to the underlying rationale through rigorous algorithmic design.
For instance, \textit{ACE} provides a more faithful implementation of the causal inference rationale by marginalizing over all possible input contexts, thereby capturing the causal necessity of each feature more accurately. 
In contrast, \textit{ICE} applies interventions within a single fixed context, only partially engaging with the causal rationale and limiting its generalizability.  
Such differences in rationale-fidelity are critical for assessing the theoretical soundness of a specific attribution method.


\section{Theoretical evaluation}
\label{Sec:theoretical evaluation and comparison}
Unlike many fields where human-annotated ground truth can serve as a benchmark, attribution explanation for DNNs lacks universally accepted ground-truth annotations. 
This makes it inherently difficult to \textit{empirically} assess the faithfulness of attribution methods. 
This limitation has become a consensus among researchers  \cite{yang2019evaluating, rao2022towards, rong2022consistent, zhou2022feature, ju2022logic, yang2019benchmarking}.
Despite numerous efforts to develop alternative empirical evaluation strategies \cite{sanchez2020evaluating, zhou2021evaluating, yue2023automatic, yang2019evaluating,yang2019bim, schulz2020restricting, gevaert2024evaluating}, 
none of strategies is widely accepted as objective.
Moreover, empirical strategies often yield inconsistent or even contradictory evaluation results, further complicating the evaluation landscape.

In light of these limitations, \textit{theoretical evaluation} has garnered increasing attention in recent years.
Beyond empirical evaluation approaches, which rely on observed behaviors or downstream performance, theoretical evaluation centers on \textit{rigorously verifying whether an attribution method adheres to formally defined faithfulness principles}. These principles are typically model-independent and dataset-independent, thereby offering a more general and principled basis.

Building upon recent work in theoretical unification, an emerging trend in the field is to conduct evaluation not only at the method level but also at the \textit{attribution family level}. 
Rather than assessing methods in isolation, these studies aim to determine whether specific families—unified by shared (re)formulation—consistently satisfy or violate core theoretical principles that define faithfulness and robustness.
This shift from instance-level to family-level evaluation enables more systematic and scalable assessments, allowing researchers to reason about properties that generalize across methods.

\subsection{\!\!\!\! Theoretical valuation of modified gradient family}
Leveraging the unified formulation introduced in  Eq.~(\ref{eqn:Modified gradient based attribution family}), prior works theoretically evaluate the faithfulness of the \textit{modified gradient attribution family}.
A fundamental principle, \textit{decision-making relevance}, was proposed by Nie et al.~\cite{nie2018theoretical} to assess whether attribution results truly reflect the decision-making process of DNNs:
\vspace{2pt}
\begin{itemize}
\item \textbf{Decision-making relevance}: A faithful attribution should highlight features relevant to model's decision process. 
\end{itemize}

Through rigorous theoretical analysis, Nie et al.~\cite{nie2018theoretical} demonstrated that two representative methods within this family, such as \textit{GBP} and \textit{Deconv}, essentially perform input recovery rather than identifying decision-relevant features. In particular, in simplified CNN settings, these methods were theoretically shown to approximately reconstruct the input, producing visually appealing yet decision-irrelevant attribution maps. Subsequent extension analyses extended to deeper and more realistic models further confirmed this behavior.

These theoretical results, further corroborated by empirical studies~\cite{salehi2021multiresolution, sixt2020explanations, wagner2019interpretable, yang2019benchmarking}, indicate that \textit{Deconv} and \textit{GBP} violate the decision-making relevance principle and fail to produce faithful explanations.

\subsection{\!\!\!\! Theoretical evaluation of feature additive family} 
\label{subsec:Evaluating feature additive attribution family}
Based on the unified formulation in Eq.~(\ref{eqn:additive}), prior work has provided a theoretical foundation for evaluating the \textit{feature additive attribution family}. 
Lundberg and Lee~\cite{lundberg2017unified} proposed three core axioms that formalize what constitutes a faithful additive explanation:
\vspace{2pt}
\begin{itemize}
\item \textbf{Local accuracy}: The attribution should precisely approximate the output for the given input.
\item \textbf{Missingness}: Features that are absent (or unobserved) should receive zero attribution.
\item \textbf{Consistency}: If a feature has a larger marginal contribution across differnt models, it should consistently receive a higher attribution score. 
\end{itemize}

Among all methods within the \textit{feature additive attribution family}, only the \textit{Shapley value} is proven to satisfy all three axioms, making it a uniquely theoretically sound choice in this family.  
This foundational result has led to its widespread adoption across diverse domains~\cite{sundararajan2020many, merrick2020explanation, chen2022explaining, kwon2022weightedshap}.

\begin{table*}[t]
\caption{Theoretical evaluations of attribution methods categorized by principle type and outcome.}
\renewcommand\arraystretch{2}
\centering
\begin{tabular}{>{\centering}p{28mm} | >{\centering}p{20mm} | >{\centering}p{28mm} | >{\centering}p{37mm}| >{\centering\arraybackslash} p{47mm}}
\toprule
\textbf{Attribution Family} & \textbf{Principle Category} & \textbf{Evaluation Principles}   & \textbf{Satisfying Methods} & \textbf{Violating Methods} \\ \hline

Modified Gradient \cite{nie2018theoretical} 
& Faithfulness
& Decision-making relevance
& others in this family 
&  \textit{Deconv}  \cite{zeiler2014visualizing}, \textit{GBP} \cite{springenberg2014striving}
\\ \hline

Feature Additive \cite{lundberg2017unified}
& Faithfulness
& Local accuracy, Missingness, Consistency
&\textit{Shapley} \cite{lundberg2017unified} 
& others in this family 
\\ \hline

Nonnegative BP \cite{sixt2020explanations}
& Faithfulness
& Output sensitivity, \\ Parameter sensitivity
& —
& All methods in this family
\\ \hline

Taylor Interaction \quad \cite{10414149, deng2021unified}
& Faithfulness
& Effect completeness, Allocation correctness, Allocation completeness
& \textit{IG} \cite{sundararajan2017axiomatic}, \textit{EG} \cite{erion2019learning},
\textit{DL Res} \cite{shrikumar2017learning}, 
 \textit{DL Rev} \cite{shrikumar2017learning}, \textit{Shapley}  \cite{lundberg2017unified}, \textit{DSHAP} \cite{lundberg2017unified}
& \textit{DTD} \cite{montavon2017explaining}, \textit{LRP-$0$/$\epsilon$} \cite{bach2015pixel}, \textit{LRP-$\alpha\beta$} \cite{bach2015pixel},
 \textit{Occ-1} \cite{zeiler2014visualizing}, \textit{Occ-p} \cite{zintgraf2017visualizing}, \textit{PDiff} \cite{zintgraf2017visualizing},
\textit{Grad$\times$Input} \cite{simonyan2013deep}, \textit{GradCAM} \cite{selvaraju2017grad}
\\ \hline

\multirow{2}{*}{\makecell[c]{Feature Removal  \\ \cite{lin2024robustness, agarwal2021towards, khan2024analyzing}}} & 
\multirow{2}{*}{\makecell[c]{Robustness}} 
& Input-perturbation  robustness
&  \multicolumn{2}{c}{
\makecell{\vspace{-3pt} \\ Context-dependent, depends  on model smoothness, \\
baseline distribution, and summary technique}
}
\\ 
\cline{3-5}
~ 
& 
~
& Model-perturbation robustness
&   \multicolumn{2}{c}{
\makecell{\vspace{-3pt} \\ Context-dependent, depends on model smoothness, \\
baseline distribution, and summary technique}
}
\\ \hline

\multirow{2}{*}{\makecell[c]{Gradient-based \\
\cite{dombrowski2019explanations, wang2020smoothed, agarwal2021towards}}} & \multirow{2}{*}{Robustness}
& Input-perturbation robustness
&  \multicolumn{2}{c}{
\makecell{\vspace{-3pt} \\ Context-dependent, depends on model smoothness}}
\\ \cline{3-5}
~ & ~
& Model-perturbation robustness
& —
& All methods in this family
\\ \hline

\bottomrule
\end{tabular}
\vspace{-5pt}
\label{tab:theoretical-evaluation-reorg}
\end{table*}

\subsection{\!\!\!\!  Theoretical evaluation of nonnegative BP family} 
\label{subsec:Evaluating nonnegative-matrix product attribution family}
Building upon the unified formulation introduced in Eq.~(\ref{eqn: Non-negative matrix product}), i.e., $\bm{a} = \prod\nolimits_{l=1}^{L} M^{(l),+} \bm{y}$, prior works turn to theoretically evaluating the \textit{non-negative BP attribution family}.
To assess the faithfulness of this family, two sensitivity principles that have been widely adopted in the literature~\cite{adebayo2018sanity, sixt2020explanations, wang2022reinforced, wang2020score, fan2020can, yeh2019fidelity} are used:
\vspace{2pt}
\begin{itemize}
\item \textbf{Output sensitivity}: A faithful attribution should be sensitive to the DNN's output. Specifically, attributions should vary significantly for different predicted categories.
\item \textbf{Parameter sensitivity}: A faithful attribution should be sensitive to the network parameters, especially those in later layers. Randomizing these parameters should substantially affect attribution results.
\end{itemize}

However, theoretical analysis by Sixt et al.~\cite{sixt2020explanations} has shown that all methods within the \textit{non-negative BP attribution family} suffer from a structural limitation: attribution results tend to converge to a nearly fixed direction, largely independent of the model's output or later-layer parameters. This convergence arises from the repeated multiplication of non-negative matrices, which acts as a form of rank-1 projection and effectively suppresses output-specific and parameter-specific information.

These theoretical findings, further corroborated by extensive empirical studies~\cite{adebayo2018sanity, sixt2020explanations, schulz2020restricting, deng2021mutual}, indicate that the \textit{non-negative BP attribution family} violates key sensitivity principles and fails to produce faithful explanations.

\subsection{\!\!\!\!  Theoretical evaluation of  Taylor interaction family} 
Building upon the unified formulation introduced in Eq.~(\ref{eqn:Taylor interaction attribution}), prior works theoretically evaluate the \textit{Taylor interaction attribution family}, which reformulates attributions as weighted sums of  independent effects and interaction effects.
To assess the reasonableness of attribution allocation within this family, three fundamental principles have been proposed for faithfulness~\cite{10414149, deng2021unified}:
\vspace{2pt}
\begin{itemize}
\item \textbf{Effect completeness}: A faithful attribution should fully account for all Taylor independent and interaction effects of the DNN, ensuring that the sum of attributions well matches the total effects.
\item \textbf{Allocation correctness}: Each effect should be assigned exclusively to the relevant variables involved, avoiding allocation to unrelated variables.
\item \textbf{Allocation completeness}: Each effect should be completely distributed among the relevant input variables without any remainder.
\end{itemize}

Theoretical analysis by Deng et al.~\cite{10414149} systematically examined fourteen methods within the \textit{Taylor interaction attribution family}. Among them, only six methods—\textit{IG, EG, DL-Res, DL-Rev, Shapley}, and \textit{Deep SHAP}—were shown to satisfy all three principles. 
In contrast, the remaining eight methods (\textit{DTD, LRP-$\epsilon$, LRP-$\alpha\beta$, Occ-1, Occ-p, PDiff, Grad$\times$Input}, and \textit{GradCAM}) were found to violate at least one principle, suggesting limited faithfulness.

These findings indicate that some methods within the \textit{Taylor interaction attribution family} achieve faithfulness under the proposed principles, while others exhibit fundamental limitations in effect allocation.
Understanding these limitations is essential for method selection.

\subsection{\!\!\!\! Theoretical evaluation of  feature removal family}
Building upon the unified formulation in Eq.~(\ref{eqn:Feature removal attribution family}), prior works have systematically investigated the  theoretical robustness of \textit{feature removal attribution family}.
In particular, two widely adopted robustness principles are employed
~\cite{lin2024robustness, khan2024analyzing, agarwal2021towards}:
\begin{itemize}
\item \textbf{Input-perturbation robustness}: Attribution results should remain stable under small input perturbations, formally quantified as $||\bm{a}(f, \bm{x}) - \bm{a}(f, \bm{x'})||_2$.
\item \textbf{Model-perturbation robustness}: Attribution results should remain stable under small model perturbations, measured as $||\bm{a}(f, \bm{x}) - \bm{a}(f', \bm{x})||_2$.
\end{itemize}

For input-perturbation robustness, several studies have shown that certain methods in the feature removal family, such as \textit{C-LIME}, \textit{Shapley}, \textit{RISE}, \textit{Occ-1}, and \textit{Occ-p}~\cite{agarwal2021towards, khan2024analyzing}, exhibit provable input-perturbation robustness under specific conditions. 
For instance, \textit{C-LIME} is provably robust when the model has bounded gradients, while  \textit{RISE} and \textit{Shapley} show robustness when the model is locally smooth. However, these guarantees are often constrained to particular settings, such as fixed baselines and certain  summary techniques.

To provide a more general and principled understanding, Lin et al.~\cite{lin2024robustness} proposed a unified theoretical framework showing that both input- and model-perturbation robustness can be bounded by three key components:
\vspace{2pt}

\noindent (1) \textit{Model smoothness $L_f$}: The model smoothness, characterized by its Lipschitz constant,  plays a central role in both robustness dimensions. A smaller Lipschitz constant usually leads to higher robustness.

\vspace{2pt}
\noindent (2) \textit{Baseline distribution $p(\bm{b}_S)$}:  The baseline distribution interacts with model smoothness to further modulate robustness, with its influence varying under input and model perturbations:

\begin{itemize}
\item For input-perturbation robustness, \textit{conditional} distribution leads to an increased robustness upper bound and weakens robustness. In contrast,  \textit{Dirac} or \textit{marginal} distributions generally exhibit stronger robustness. 

\item For model-perturbation robustness, \textit{conditional} distribution enhances robustness by limiting model perturbations within a specific subdomain $\mathcal{X}$. In contrast, \textit{Dirac} and \textit{marginal} distributions do not impose such restrictions, leading to weaker robustness.
\end{itemize}

\vspace{2pt}
\noindent (3) \textit{Summary technique $\psi(\cdot)$}: The summary technique adopted by each method also plays an important role. Approaches using \textit{leave-one-out}, \textit{Shapley}, or \textit{Banzhaf} summaries
generally exhibit both weaker input-perturbation and model-perturbation robustness, compared to those using aggregation schemes like \textit{mean when included}, as adopted in \textit{RISE}.

In summary, the input and model robustness of feature removal attribution methods is governed by a triad of interacting factors—model inherent  smoothness, baseline distribution, and summary technique.

\subsection{\!\!\! Theoretical evaluation of  gradient based family}
The input-perturbation and model-perturbation robustness principles described above are also employed to evaluate the \textit{gradient-based attribution family}, which encompasses both the modified gradient methods and the Gradient$\times$Input methods.
\vspace{2pt}

\noindent \textbf{(1) Input-Perturbation Robustness.}
Numerous studies have shown that gradient-based attribution methods, including \textit{Gradient}, \textit{Grad$\times$Input}, \textit{IG}, and \textit{GradCAM}, are highly sensitive to small changes in input~\cite{kindermans2019reliability, ghorbani2019interpretation, dombrowski2019explanations, viering2019manipulate}.
This sensitivity often leads to substantially different attribution results for similar inputs, challenging their reliability.

Theoretically, this instability has been attributed to model smoothness.
Smoother models show stronger robustness under input perturbations. 
Key smoothness metrics include principal curvatures \cite{dombrowski2019explanations}, the Frobenius norm of Hessian matrix \cite{dombrowski2022towards}, and Lipschitz constant \cite{wang2020smoothed}. 
Several approaches have been proposed to mitigate this robustness issue, such as using smoother activation functions, applying  weight decay or Hessian  regularization. 
Additionally, \cite{agarwal2021towards} shows that \textit{smooth gradients}, a method based on stochastic smoothing, is provably robust when the model has a bounded gradient.
\vspace{2pt}

\noindent \textbf{(2) Model-Perturbation Robustness.}
Gradient-based attribution methods, such as \textit{Gradient} and \textit{Grad-CAM},  are also vulnerable to manipulation via model parameter perturbations. Heo et al.~\cite{heo2019fooling} empirically showed that it is possible to modify a model such that its predictions remain unchanged, yet its attribution explanations are arbitrarily altered.

Further theoretical work~\cite{anders2020fairwashing} supports this observation, proving that for any given model $f$, there exists an alternative model $\tilde f$ with identical model outputs but substantially different gradient-based attributions. 
This indicates a fundamental lack of model robustness for this family.

\subsection{Our Insight: Lessons from theoretical evaluation} 
\label{subsec:our perspective3}
In this section, we reflect on the role and limitations of existing theoretical evaluations, and offer insights on how to interpret and effectively leverage existing works in this area.
\vspace{4pt}

\noindent \textbf{(1) Existing evaluations focus on falsifiability; verifiability remains elusive.}
One often overlooked point is that current faithfulness principles predominantly serve as falsification tools—that is, they offer \textit{necessary but not sufficient conditions} for evaluating attribution quality. 
In practice, principles such as output sensitivity and parameter sensitivity serve as falsification-oriented sanity checks, aiming to identify attribution methods that exhibit pathological or arbitrary behavior.

While falsifiability is widely regarded as a hallmark of scientific rigor, it does not imply verifiability.
Even if an attribution method satisfies all known falsifiability principles, this offers no guarantee of faithfulness, as these principles capture only limited facets of attribution behavior and may still permit spurious or misleading explanations.
This asymmetry underscores a fundamental challenge in attribution research: falsification is operationally feasible, whereas verification remains far more elusive and theoretically unresolved.

To date, no existing principle or evaluation framework provides a universally applicable \textit{sufficient condition} to verify attribution faithfulness across diverse models, architectures, and tasks.
This limitation underscores the need for future research to move beyond falsifiability and advance toward verifiable guarantees of attribution quality.
\vspace{4pt}

\noindent  
\textbf{(2) Existing evaluations primarily serve to eliminate unfaithful methods.}
Given the fundamental asymmetry between falsifiability and verifiability, current faithfulness evaluations are best understood as tools for identifying and ruling out unfaithful attribution methods.  
While satisfying these principles does not guarantee a method is truly faithful, violating them often provides strong evidence of unfaithfulness.  
To this end, we explicitly summarize methods that violate one or more faithfulness principles (see Table~\ref{tab:violating faithfulness}), and advise particular caution when applying these methods in practice, especially in those high-stakes or scientifically critical applications.

\begin{table}[t]
\caption{Attribution methods violating faithfulness principles.} 
\label{tab:violating faithfulness}
\renewcommand\arraystretch{1.8}
\centering
\begin{tabular}{p{26mm} | p{54mm}}
\toprule
\textbf{Attribution methods} & \textbf{Violated faithfulness principles} \\
\hline
\textit{Occ-1/Occ-p/PDiff} & Allocation Fidelity \\ \hline
\textit{Grad$\times$Input/GradCAM} & Allocation Fidelity \\ \hline
\textit{DL-Res/LIME} & Axiomatic Fidelity \\ \hline
\textit{RectG/ExBP} & Output/Parameter Sensitivity \\ \hline
\textit{Deconv/GBP} & Decision Relevance, Output/Param. Sensitivity \\ \hline
\textit{LRP-$\alpha\beta$/DTD} & Output/Param. Sensitivity,  Allocation Fidelity \\ \hline
\textit{LRP-$\epsilon$} & Axiomatic Fidelity,  Allocation Fidelity \\ \hline
\bottomrule
\end{tabular}
\vspace{-6pt}
\end{table}

\section{Takeaways on practical guidance}
\label{sec:Expected benefits}
Beyond advancing theoretical understanding, it is equally important to translate these insights into actionable guidance for real-world use. 
This section aims to bridge the gap between theory and practice by highlighting how theoretical findings can inform both the use and development of attribution methods. 
Specifically, we focus on:  
(i) providing guidance for informed method selection and usage;  
and (ii) inspiring the design of novel attribution techniques and evaluation strategies grounded in principled theoretical foundations.

 
\subsection{Theoretical guidance for method selection and usage}
This survey offers practical guidance for selecting and applying attribution methods in real-world scenarios—a particularly valuable contribution given the well-known challenges in empirically evaluating these methods.
\vspace{4pt}

\noindent
\textbf{(1) Favoring theoretically principled methods.}  
Attribution methods tend to be more reliable and trustworthy in practice when they (i) exhibit compatibility with multiple reformulation families, (ii) are underpinned by strong theoretical rationales, and (iii) satisfy established faithfulness principles.

A prominent example is \textit{Shapley value}, which is compatible with as many as five reformulation families, demonstrating its broad applicability across different paradigms.  
In addition, it draws support from multiple foundational rationales, including surrogate approximation, information theory, causal inference, and interaction effect allocation. 
Furthermore, \textit{Shapley value}  satisfies all major faithfulness principles, further reinforcing its theoretical soundness.  
These multi-level strengths have made Shapley-based methods a widely preferred choice in explainability research and applied settings.
\vspace{4pt}

\noindent
\textbf{(2) Exercising caution with weakly supported methods.}  
Attribution methods based on incomplete theoretical justifications or those known to violate core faithfulness principles may yield unreliable or even misleading explanations in practice.

For example, although the \textit{DTD} method has seen wide adoption, theoretical analyses reveal that it suffers from both output and parameter insensitivity (see Section~\ref{subsec:Evaluating nonnegative-matrix product attribution family}).  
Moreover, the validity of its foundational rationale—deep Taylor decomposition—remain contested (see Section~\ref{subsec:our perspective2}).  
Thus, practitioners should be fully aware of these limitations before applying \textit{DTD} explanations in sensitive or high-stakes applications.
\vspace{4pt}

\noindent 
\textbf{(3) Ensuring proper implementation practices.}  
When applying attribution methods, it is crucial to consider not only their theoretical soundness but also implementation details that may affect stability and reliability.

Take \textit{LIME} as an example: despite being widely adopted, its performance can be unstable due to randomness in the surrogate sampling process.  
If not properly controlled, this instability can compromise the quality of local approximations and thereby undermine the faithfulness of the explanation (see Section~\ref{subsec:our perspective2}).  
To ensure effective application, it is critical to implement stabilization techniques or use improved variants such as \textit{BayLIME} or \textit{OptiLIME}.

\subsection{Theoretical guidance for method and evaluation design} 
Beyond method selection and usage, this work offers valuable insights for guiding the design of new attribution methods and new theoretical evaluation strategies.
\vspace{4pt}

\noindent
\textbf{(1) Guiding the design of new attribution methods}.   
This survey identifies which types of attribution methods hold theoretical promise and which exhibit intrinsic limitations.  
For example, methods in the \textit{nonnegative BP attribution family} may be deprioritized in future work, as their attributions are largely independent of both network outputs and parameters, limiting their practical relevance.  
In contrast, \textit{Shapley-based methods} appear to warrant more attention and become a particularly promising avenue for continued exploration, given their strong theoretical foundation and broad applicability.
\vspace{4pt}

\noindent
\textbf{(2) Guiding the design of  new theoretical evaluations}.  
This work emphasizes the value of conducting theoretical evaluations at the level of attribution families, rather than evaluating methods in isolation.  
By organizing attribution methods into unified families based on shared reformulations, one can uncover common structural behaviors that naturally suggest which theoretical principles are most appropriate for evaluation. 
Such a strategy enables more coherent and scalable evaluation by assessing multiple methods together under a shared theoretical lens (see Section~\ref{Sec:theoretical evaluation and comparison}).

For instance, the \textit{non-negative BP family}, identified through reformulation unification, exhibits a shared structural behavior: its products of non-negative BP matrices tend to converge to a rank-1 matrix as network depth increases. 
This convergence results in a loss of sensitivity to both model outputs and parameters, a limitation that typically remains hidden when methods are analyzed in isolation. 
Recognizing this structural limitation, two targeted faithfulness principles,  \textit{Output Sensitivity} and \textit{Parameter Sensitivity}, are conducted to systematically assess the reliability of methods within this family (see Section~\ref{subsec:Evaluating nonnegative-matrix product attribution family}).
This example highlights the value of family-level evaluation, where shared structures enable more targeted and interpretable evaluations.

\section{Future work}
\label{sec:Future work}
In this section, we outline key future directions for theoretical research on attribution explanations. 

\subsection{Proposing theoretically reliable attribution methods}  
Although certain  attribution methods, such as \textit{Shapley value}, have been shown grounded in relatively sound theoretical foundations, they are not without limitations \cite{kwon2022weightedshap, kumar2020problems, heskes2020causal}. Therefore, developing attribution methods with stronger theoretical guarantees remains a central research focus.

One promising strategy is to refine and extend attribution methods that are already grounded in strong theoretical foundations.
For instance, although the \textit{Shapley value} enjoys strong theoretical underpinnings, it still faces concerns  regarding its theoretical rigor.
Its standard formulation has been shown to inadequately capture causal relationships, 
prompting the development of causally informed variants such as \textit{Causal Shapley} \cite{heskes2020causal} and \textit{Rational Shapley} \cite{watson2022rational}.  
Additionally, classic Shapley methods primarily focus on attributing importance to individual input variables, and often overlook higher-order interactions among features.
To address this limitation, a growing body of work have  developed interaction-based attribution approaches, which offer more fine-grained and structurally comprehensive explanations \cite{sundararajan2020shapley, sikdar2021integrated, janizek2021explaining, tsai2023faith, ren2023defining}.

\subsection{Developing more comprehensive theoretical evaluations} 
Current theoretical evaluations are typically confined to a limited number of attribution families, and comparisons are often restricted to methods within each individual family.
As a result, a substantial number of methods remain outside the scope of existing theoretical evaluation frameworks.
This narrow scope limits our ability to form a comprehensive evaluation of attribution faithfulness, leaving critical blind spots in both theoretical analysis and practical deployment.

To address this gap, future efforts should systematically expand theoretical evaluation coverage.  
First, this requires incorporating under-evaluated attribution families and developing faithfulness principles tailored to their unique assumptions and mechanisms.
For example, surrogate-based methods  may require criteria centered on approximation fidelity, such as bounds on surrogate model errors.  
Second, there is a growing need for cross-family evaluation metrics that support comparison across fundamentally different attribution paradigms.

\subsection{Beyond falsifiability: toward verifiable evaluation}
While there is broad consensus that attribution methods should faithfully reflect model decision logic, the notion of "faithfulness" remains under-specified and lacks a universally accepted formalization.    
Most existing theoretical evaluations primarily establish \textit{necessary conditions} for falsifying faithfulness, such as output sensitivity and parameter sensitivity, which help detect clearly unfaithful methods. 
However, \textit{sufficient conditions} for fully verifying faithfulness remain elusive.

To move beyond falsifiability and advance towards verifiability, future work must aim to establish operational and generalizable definitions of faithfulness, alongside rigorous sufficient conditions for verification.
A few studies have attempted to formalize the notion of faithfulness from different perspectives  ~\cite{yeh2019fidelity, subhash2022makes, potyka2022towards, azzolin2024perks}, but none have yet achieved consensus or wide adoption. 
Additionally, some researchers argue that in the context of DNNs, calculating the contribution of individual input variables while ensuring full faithfulness is a challenging task. 
As a result, the focus has shifted towards interaction-based attribution methods, which assign importance to cooperative subsets of variables, rather than isolating individual input variables \cite{ren2023defining}.


\subsection{Toward context-aware attribution}  
Future research should place greater emphasis on tailoring attribution methods to specific models and application domains, as different architectures and use cases often entail distinct interpretability requirements.

From a model perspective, different network architectures may require different attribution strategies.  
For example, in CNNs, methods such as \textit{Grad-CAM} are widely adopted due to their ability to highlight spatially localized, class-relevant regions.
However, these methods are not directly applicable to Transformer-based architectures or large language models (LLMs), which lack explicit spatial hierarchies.
In such cases, attribution strategies based on attention mechanisms or Shapley value are often more appropriate, as they better align with the internal structure and representation of these models.

From a domain perspective, different application areas prioritize different explanation goals and faithfulness criteria.
In high-stakes domains such as AI for Science (AI4S), attribution methods are not only expected to explain predictions, but also to facilitate scientific discovery.
Consequently, attribution approaches like the influential subset attribution family have gained prominence. These methods identify functionally important substructures in molecular graphs or protein structures, helping to uncover chemically meaningful components such as functional groups, active binding sites, or reactive centers \cite{wu2023chemistry}.

In summary, the effectiveness of an attribution method is inherently context-dependent.
Future research should move toward context-aware attribution, prioritizing both model-specific and domain-specific selection strategies to ensure that interpretability tools are well aligned with architectural properties and application goals.

\bibliography{survey}
\bibliographystyle{IEEEtran}



\ifCLASSOPTIONcaptionsoff
  \newpage
\fi

\vspace{-20pt}
\begin{IEEEbiography}[{\includegraphics[width=1in,height=1.5in,clip,keepaspectratio]{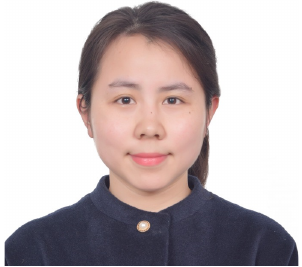}}]{Dr. Huiqi Deng} is an assistant professor at Xi'an Jiao Tong University, China. 
She received her Ph.D. degree in applied mathematics from Sun Yat-sen University, China, in 2021.
She was a visiting scholar at the Texas A\&M University (TAMU) and Hong Kong Baptist University (HKBU). 
Her research focuses on explainable AI and various aspects of trustworthy AI, such as generalization and robustness.
To date, she has published over 20 papers in leading academic journals and conferences, such as IEEE TPAMI, Pattern Recognition, NeurIPS, ICML, NeurIPS, ICLR, AAAI, and KDD.
\end{IEEEbiography}

\vspace{-20pt}

\begin{IEEEbiography}[{\includegraphics[width=1in,height=1.5in,clip,keepaspectratio]{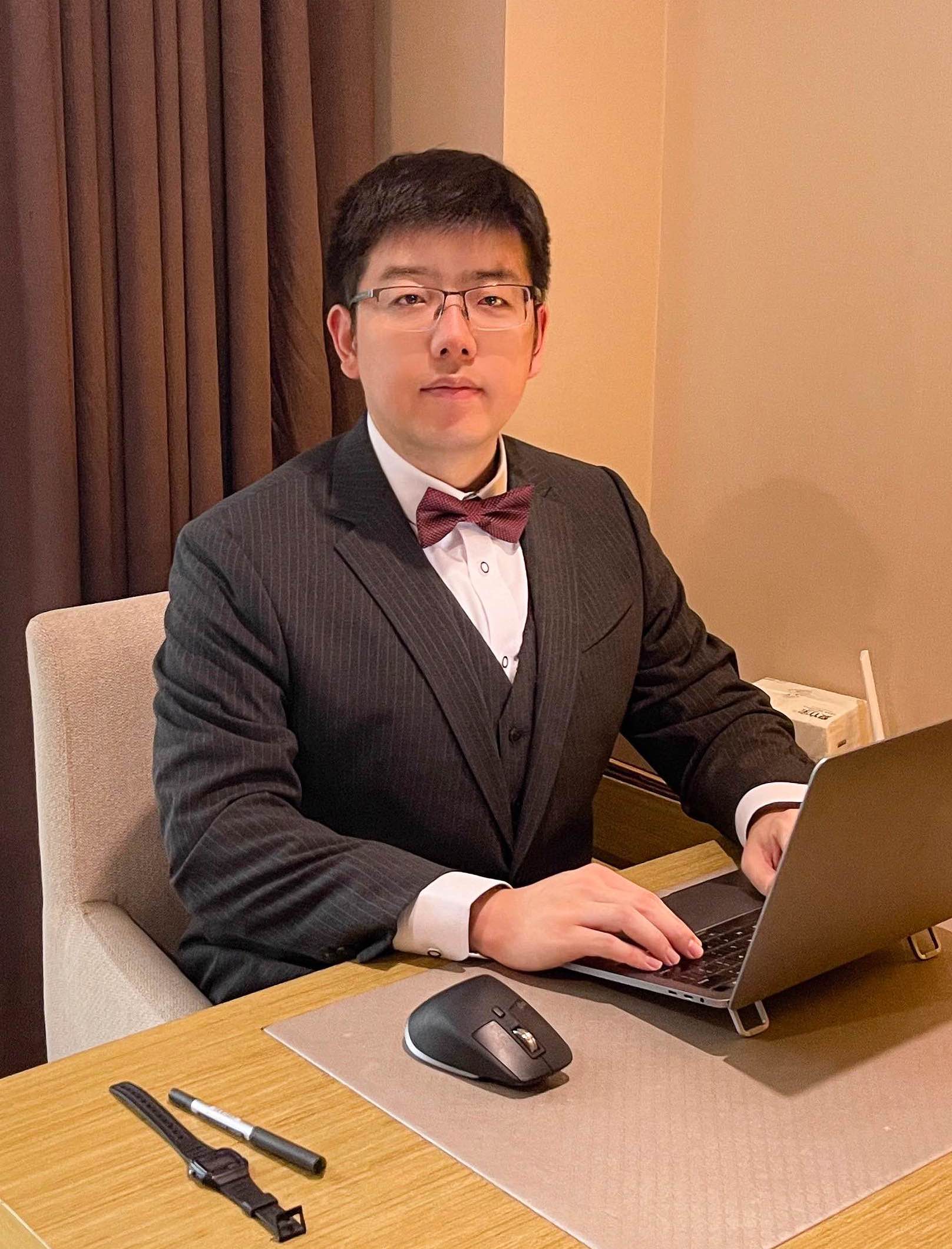}}]
{Dr. Hongbin Pei}  is an Assistant Professor at Xi’an Jiaotong University, China. He received his B.S., M.S., and Ph.D. degrees from Jilin University in 2012, 2015, and 2021, respectively. He was a visiting scholar at the University of Illinois at Urbana-Champaign (UIUC) and Hong Kong Baptist University (HKBU). His research focuses on graph learning, geometric deep learning, and spatio-temporal data mining, with applications for social good. He serves as a senior program committee member and reviewer for conferences and journals, including CVPR, ICML, ICLR, IEEE TPAMI, and TNNLS.
\end{IEEEbiography}

\vspace{-20pt}
\begin{IEEEbiography}
[{\includegraphics[width=1in,height=1.5in,clip,keepaspectratio]{./figures/zhang.pdf}}]
{Dr. Quanshi Zhang} is an associate professor at Shanghai Jiao Tong University, China.
He received Ph.D. degree from the University of Tokyo in 2014. From 2014 to 2018,
he was a post-doctoral researcher at the
University of California, Los Angeles. His research interests include machine learning and computer vision. In particular, he
has made influential research in explainable
AI (XAI). He won the ACM China Rising
Star Award at ACM TURC 2021. He is the
speaker of the tutorials on XAI at IJCAI 2020 and IJCAI 2021. He
was the co-chairs of the workshops towards XAI in ICML 2021, AAAI
2019, and CVPR 2019.
\end{IEEEbiography}

\vspace{-20pt}
\begin{IEEEbiography}[{\includegraphics[width=1in,height=1.5in,clip,keepaspectratio]{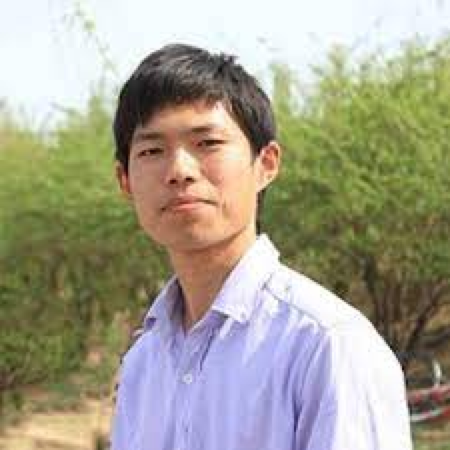}}]{Dr. Mengnan Du} is an Assistant Professor of Data Science at New Jersey Institute of Technology. He earned his Ph.D. in Computer Science from Texas A\&M University. His research interests lie within the broad domain of trustworthy machine learning, with a particular emphasis on its intersection with large language models (LLMs). He has published over 90 papers in top-tier conferences including NeurIPS, ICLR, and ICML, as well as prestigious journals such as TPAMI, CACM, and Cell Patterns. His work has garnered more than 7,800 Google Scholar citations. He actively contributes to the academic community by organizing workshops and tutorials at major conferences including AAAI-24, WWW-24, and COLM-25. He serves as the Senior Area Chair for EMNLP-25, Area Chairs for prestigious conferences including NeurIPS-25, ICML-25, ACL-25, and AISTATS-25, and as an Associate Editor for Applied AI Letters.
\end{IEEEbiography}
\vfill

\end{document}